\documentclass[11pt]{article}

\usepackage[final]{acl}

\usepackage{times}
\usepackage{latexsym}
\usepackage{enumitem}
\usepackage{multirow}
\usepackage{amsmath}
\usepackage{amssymb}
\usepackage{booktabs}
\usepackage[linesnumbered,ruled,vlined]{algorithm2e}
\usepackage{amsmath}
\usepackage{tabularx}
\usepackage{xcolor}
\usepackage[table]{xcolor}
\usepackage[T1]{fontenc}

\usepackage[utf8]{inputenc}

\usepackage{microtype}

\usepackage{inconsolata}

\usepackage{graphicx}

\usepackage[table,dvipsnames]{xcolor}
\usepackage{booktabs}
\usepackage{amsmath}
\usepackage{amssymb}
\usepackage{geometry} 

\newcommand{\stepRewrite}[1]{\noindent\textbf{\textcolor{RoyalBlue}{\textsc{[Rewrite]}}} \textcolor{RoyalBlue}{#1}}
\newcommand{\stepPrune}[1]{\noindent\textbf{\textcolor{BrickRed}{\textsc{[Prune]}}} \textcolor{BrickRed}{#1}}
\newcommand{\stepFuse}[1]{\noindent\textbf{\textcolor{BurntOrange}{\textsc{[Fuse]}}} \textcolor{BurntOrange}{#1}}
\newcommand{\stepKeep}[1]{\noindent\textbf{\textcolor{ForestGreen}{\textsc{[Keep]}}} \textcolor{ForestGreen}{#1}}

\title{CRISP: Compressing Redundancy in Chain-of-Thought via Intrinsic Saliency Pruning}

\author{Yangsong Lan, Hongliang Dai, Piji Li$^{\ast}$ \\
\textsuperscript{\rm 1} College of  Artificial Intelligence, \\
Nanjing University of Aeronautics and Astronautics, Nanjing, China\\
\textsuperscript{\rm 2} The Key Laboratory of Brain-Machine Intelligence Technology, Ministry of Education, Nanjing, China.\\
  \texttt{\{lys2962331781, hongldai, pjli\}@nuaa.edu.cn} \\}

\begin{document}
\maketitle
\renewcommand{\thefootnote}{\fnsymbol{footnote}}
\footnotetext[1]{Corresponding author.}
\renewcommand{\thefootnote}{\arabic{footnote}}
\begin{abstract}
Long Chain-of-Thought (CoT) reasoning is pivotal for the success of recent reasoning models but suffers from high computational overhead and latency. While prior works attempt to compress CoT via external compressor, they often fail to align with the model's internal reasoning dynamics, resulting in the loss of critical logical steps. This paper presents \textbf{C}ompressing \textbf{R}edundancy in Chain-of-Thought via \textbf{I}ntrinsic \textbf{S}aliency \textbf{P}runing (\textbf{CRISP}), a framework that compresses CoT by exploiting the model's intrinsic saliency. Our analysis reveals a distinct phenomenon: the reasoning termination token \texttt{</think>} acts as an information anchor, where its attention pattern effectively demarcates essential reasoning from redundancy. Based on this finding, we design a policy that utilizes these intrinsic attention signals to guide atomic compression operations. In contrast to coarse-grained pruning strategies, CRISP strategically distills the reasoning chain to maximize information density while preserving logical coherence. Empirical results across various backbone models and mathematical datasets demonstrate that CRISP achieves a 50-60\% reduction in token count without compromising accuracy, effectively mitigating the efficiency bottleneck of long-context reasoning. We open-source our implementation to facilitate further research in efficient reasoning\footnote{Our implementation is available at \href{https://github.com/yslanprime/CRISP}{GitHub}.}.

\end{abstract}

\section{Introduction}

The emergence of reasoning-oriented LLMs, represented by OpenAI o1~\cite{jaech2024openai}, DeepSeek-R1~\cite{guo2025deepseek}, QwQ~\cite{renze2024benefits}, and Kimi k1.5~\cite{team2025kimi}, marks a significant paradigm shift, demonstrating superior performance in complex reasoning domains. While these models achieve superior reasoning capabilities by decomposing complex problems into extensive Chains-of-Thought and employing iterative verification~\cite{wei2022chain}, this approach incurs substantial computational overhead~\cite{feng2025efficient,sui2025stop}. Such inefficiency renders deployment infeasible in resource-constrained environments, making it imperative to develop methods that distill compact reasoning paths without compromising model fidelity.

\begin{figure}[t] 
    \centering
    \includegraphics[width=\linewidth]{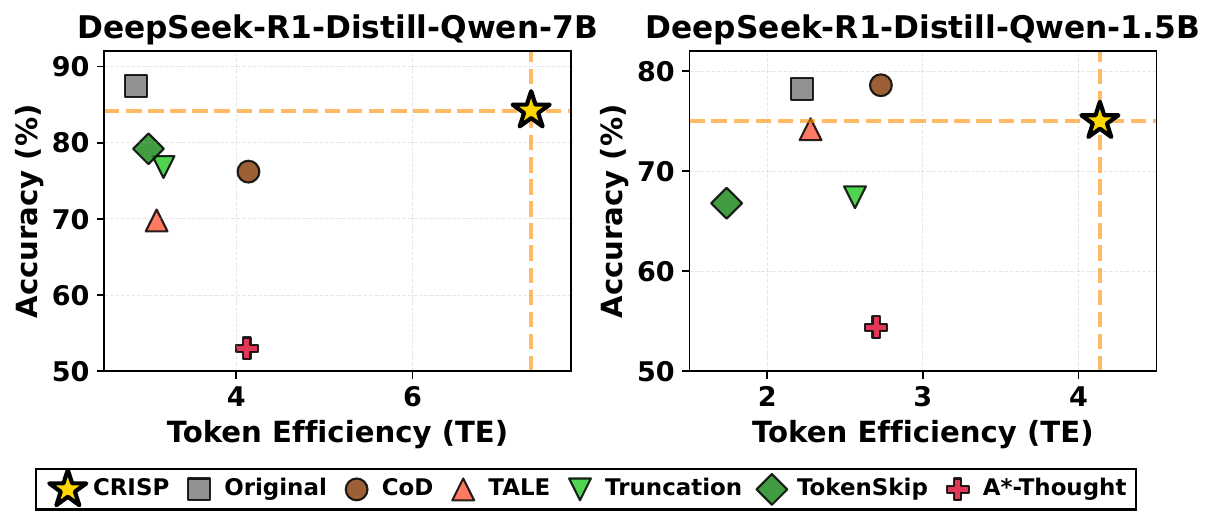}
    \caption{\textbf{Token Efficiency (TE.) vs. Accuracy on DeepSeek-R1-Distill-Qwen-7B/1.5B}. CRISP achieves the best trade-off, significantly outperforming baselines in efficiency while maintaining high accuracy.}
    \label{fig:efficiency_comparsion}
    \vspace{-8pt}
\end{figure}

To alleviate this computational burden, prevalent approaches~\cite{xia2025tokenskip,xu2025thought,yan2025long} resort to CoT compression, typically utilizing external proxy models to prune redundancy. However, this reliance on external compressors introduces a fundamental misalignment: these proxies are agnostic to the source model's intrinsic reasoning dynamics. External proxies often misclassify essential intermediate steps, particularly self-corrections, as redundancy, thereby ignoring their critical role in the source model’s logical continuity. Consequently, fine-tuning on such mutilated sequences risks disrupting the coherence of the reasoning chain. This limitation motivates two pivotal research questions: (1) How can we identify pivotal reasoning steps solely based on the model’s intrinsic signals? (2) How can we synthesize a concise yet coherent reasoning chain utilizing these identified pivots?

In this work, we address these challenges by investigating the internal attention mechanisms of reasoning models. We observe that the \texttt{</think>} token, serving as the delimiter of the reasoning phase, acts as a critical information anchor. Our analysis reveals that during the generation of the final answer, the model attends minimally to intermediate reasoning steps but maintains high attention weights on the \texttt{</think>} token. We empirically demonstrate that the attention pattern at this position reliably indicates the saliency of preceding reasoning steps, effectively distinguishing essential logic from redundancy.

Building upon this observation, we propose \textbf{CRISP}, a framework that exploits the attention landscape at the \texttt{</think>} token to steer the compression process. We cast CoT compression as a search problem defined over a quadruplet of atomic operators: \textsc{Fuse}, \textsc{Prune}, \textsc{Rewrite}, and \textsc{Keep}. By employing a reward function that harmonizes step-wise saliency with sequence length, our method efficiently navigates the reasoning space. To mitigate the logical discontinuities often introduced by discrete search operations, we further employ an advanced LLM-based refiner to refine the retrieved paths. This step effectively restores semantic coherence, yielding compressed chains that are both concise and logically fluid. Subsequently, we fine-tune the target model on these refined sequences via a multi-task learning objective.

As shown in Figure~\ref{fig:efficiency_comparsion}, CRISP achieves a superior efficiency-accuracy trade-off over strong baselines. Our framework effectively isolates essential reasoning pivots from redundancy, maintaining robust capabilities even under strict constraints.

To summarize, our main contributions are as follows:
\begin{itemize}[leftmargin=*, nosep]
    \item We identify that the attention weights at the \texttt{</think>} token serve as a reliable intrinsic indicator of reasoning step saliency, effectively distinguishing essential logic from redundancy without external proxies.
    \item We propose \textbf{CRISP}, a framework that optimizes CoT via a greedy search over four atomic operators (\textsc{Fuse}, \textsc{Prune}, \textsc{Rewrite}, \textsc{Keep}) guided by intrinsic signals.
    \item We demonstrate that \textbf{CRISP} offers a superior trade-off between efficiency and accuracy, significantly reducing inference overhead without compromising the backbone model's performance.
\end{itemize}

\section{Related Works}
\label{sec:related_works}
\paragraph{Reasoning in Large Language Models}
Chain-of-Thought has established itself as a cornerstone paradigm for Large Language Models~\cite{wangr4,wei2022chain}, enhancing interpretability and accuracy by decomposing complex tasks into intermediate reasoning steps~\cite{wang2022self,zhou2022least}. Subsequent frameworks, such as Tree of Thoughts (ToT)~\cite{yao2023tree} and Program of Thoughts (PoT)~\cite{chen2022program}, have further expanded these capabilities. More recently, models like DeepSeek-R1~\cite{guo2025deepseek}, Kimi k1.5~\cite{team2025kimi} have demonstrated that reinforcement learning algorithms~\cite{shao2024deepseekmath,yu2025dapo,zheng2025group}, coupled with meticulously designed reward signals, can unlock even deeper reasoning potential. However, this enhanced reasoning capability typically comes at the cost of excessive token generation and significant computational overhead~\cite{sui2025stop,chiang2024over}. Consequently, a growing body of research is shifting focus toward efficient reasoning, aiming to compress the CoT process and achieve an optimal trade-off between model performance and inference latency.

\begin{figure*}[t]  
    \centering
    \includegraphics[width=\textwidth]{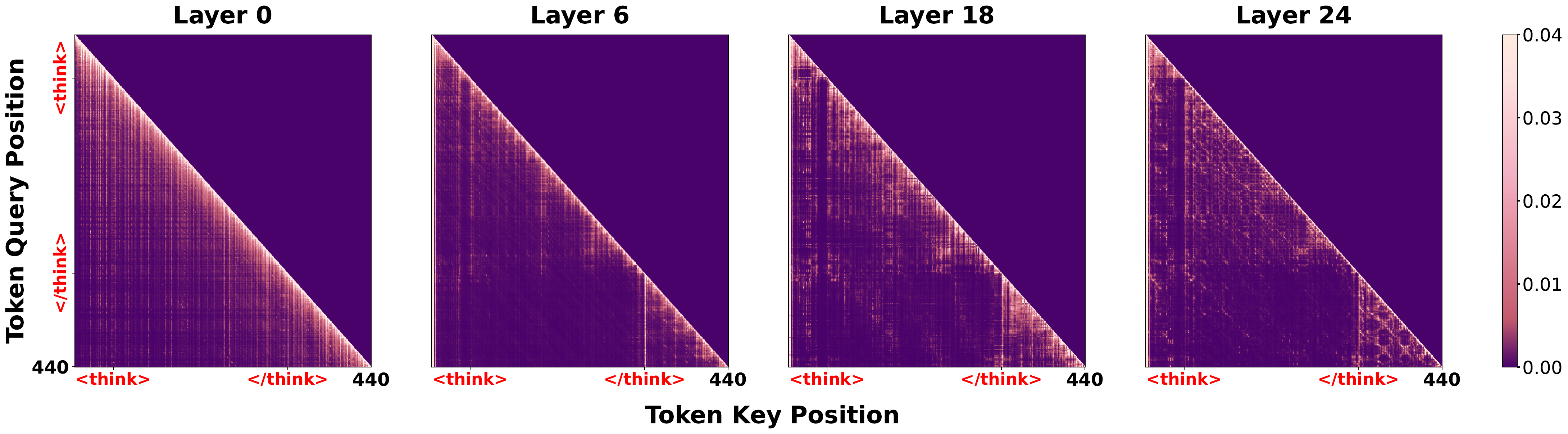} 
    \caption{\textbf{Visualization of layer-wise attention dynamics in DeepSeek-R1-Distill-Qwen7B.}The heatmaps depict layer-wise attention distributions during inference. While shallow layers exhibit uniform attention across the context, deep layers reveal the \texttt{</think>} token functioning as a semantic anchor, progressively aggregating information from the reasoning chain to guide final answer generation.}
    \label{fig:attention_multi_layers}
     \vspace{-8pt}
\end{figure*}

\paragraph{Chain-of-Thought Compression}
To mitigate inference latency, recent research seeks to compress CoT outputs without compromising reasoning efficacy~\cite{cui2025stepwise,wang2025r1,qiao2025concise}. Initial efforts utilizing prompt engineering strategies~\cite{xu2025chain,han2025token} attempt to constrain sequence length via explicit instructions, yet often struggle with granular control and adherence to constraints, leading to potential degradation in generation quality. Reinforcement learning frameworks ~\cite{aggarwal2025l1,luo2025o1} explicitly incentivize brevity by incorporating length penalties into the reward function. However, this paradigm entails significant computational overhead and exhibits acute sensitivity to reward shaping, often leading to optimization instability. Concurrently, supervised fine-tuning paradigms~\cite{xia2025tokenskip,yan2025long,xu2025thought} aim to distill concise reasoning by pruning redundant steps. However, these methods typically utilize auxiliary models as external compressors to condense trajectories. This dependency creates a misalignment, as the external compression logic often diverges from the target model’s intrinsic generation dynamics. Distinguishing itself from approaches that depend on extrinsic constraints or offline heuristics, our method leverages the model’s intrinsic attention mechanisms to guide adaptive CoT compression, ensuring that efficiency is derived directly from the instance-specific reasoning process.

\section{\texttt{</think>} as the Information Anchor}
\label{sec:information_anchor}
Unlike prior approaches that rely on external compressors to condense CoT, our objective is to determine whether the model intrinsically distinguishes the contribution of specific steps toward the final answer. To investigate the inherent attention patterns governing this interaction, we visualize the layer-wise attention maps of DeepSeek-R1-Distill-Qwen7B on GSM8K samples, as illustrated in Figure \ref{fig:attention_multi_layers}. Motivated by the hypothesis that specific tokens function as informational anchors~\citep{wang2023label, li2025thinkless}, our analysis focuses on the temporal dynamics of the \texttt{<think>} and \texttt{</think>} tokens, which delimit the reasoning boundaries.

Our visualization reveals that while attention distribution remains relatively uniform in early layers, the \texttt{</think>} token progressively functions as a semantic anchor in deeper layers, aggregating information from the preceding reasoning chain. Crucially, during the generation of the final answer, the model predominantly attends to the representation at the \texttt{</think>} position, while direct attention to the raw reasoning chain diminishes. Consistent with findings in ~\citep{choi2025think}, steps that make high contributions to the solution retain high attention scores specifically within the \texttt{</think>} token’s attention column, suggesting that the model compresses the reasoning history into this single token state. \textbf{Full attention profiles across all layers for both the 1.5B and 7B scales are detailed in Appendix \ref{sec:appendix_attention}.}

\begin{figure}[t] 
    \centering
    \includegraphics[width=\linewidth]{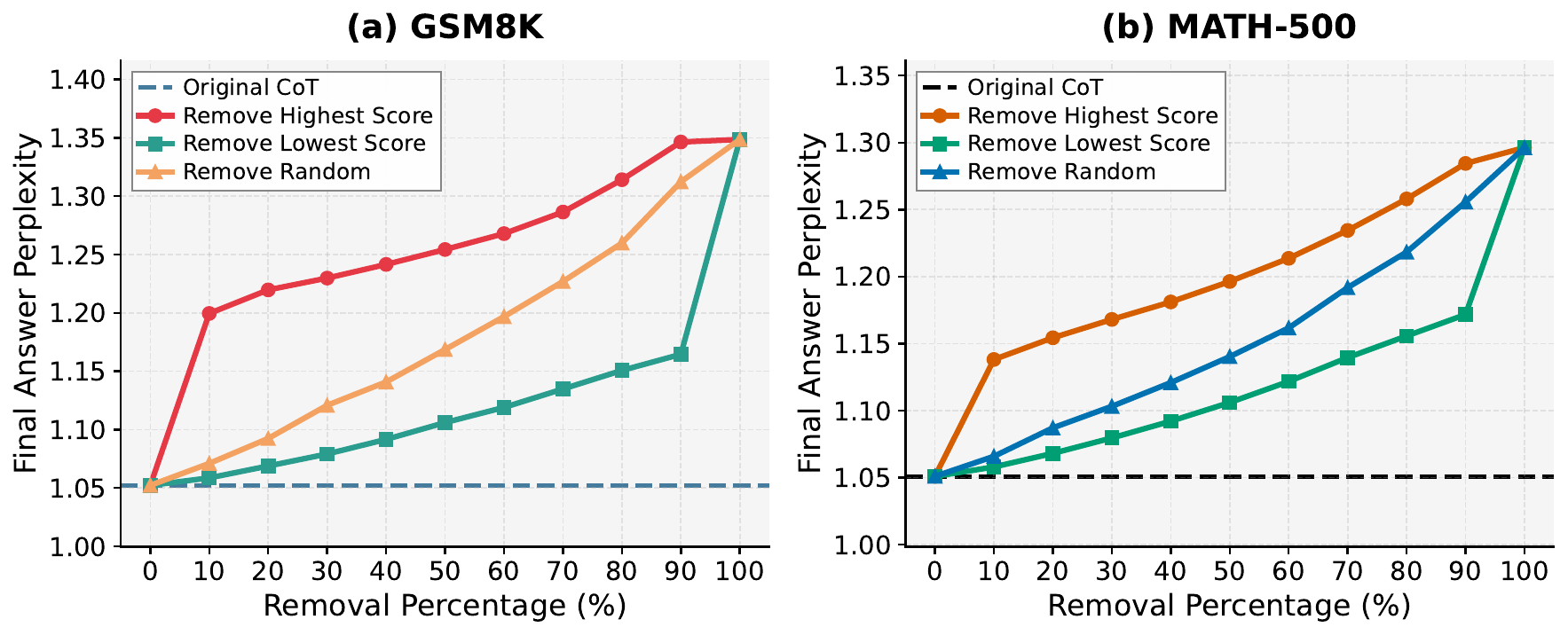}
    \caption{\textbf{Validation of anchor-guided redundancy identification.} Pruning reasoning steps with high attention to the </think> anchor precipitates a sharp PPL spike, whereas removing low-attention steps results in a significantly more gradual increase.}
    \label{fig:ppl_removal_analysis}
    \vspace{-1em}
\end{figure}

To empirically validate the role of \texttt{</think>} as a proxy for identifying information redundancy, we conducted a stepwise pruning experiment on the GSM8K and MATH-500 datasets, as detailed in Figure \ref{fig:ppl_removal_analysis}. We selectively pruned reasoning steps corresponding to the lowest, highest, and random attention scores registered at the \texttt{</think>} anchor, measuring the subsequent impact on the Perplexity (PPL)~\cite{jelinek1977perplexity} of the final answer. Our empirical results demonstrate that pruning steps with high attention scores induces a sharp spike in PPL, suggesting these steps encode critical information. Conversely, pruning low-attention steps leads to only a marginal PPL increase, while random pruning results in an intermediate performance degradation. This empirical insight serves as the foundational premise for our proposed framework, \textbf{CRISP}, which leverages these intrinsic attention signals to guide efficient and adaptive chain-of-thought compression.

\begin{figure*}[t]  
    \centering
    \includegraphics[width=\textwidth]{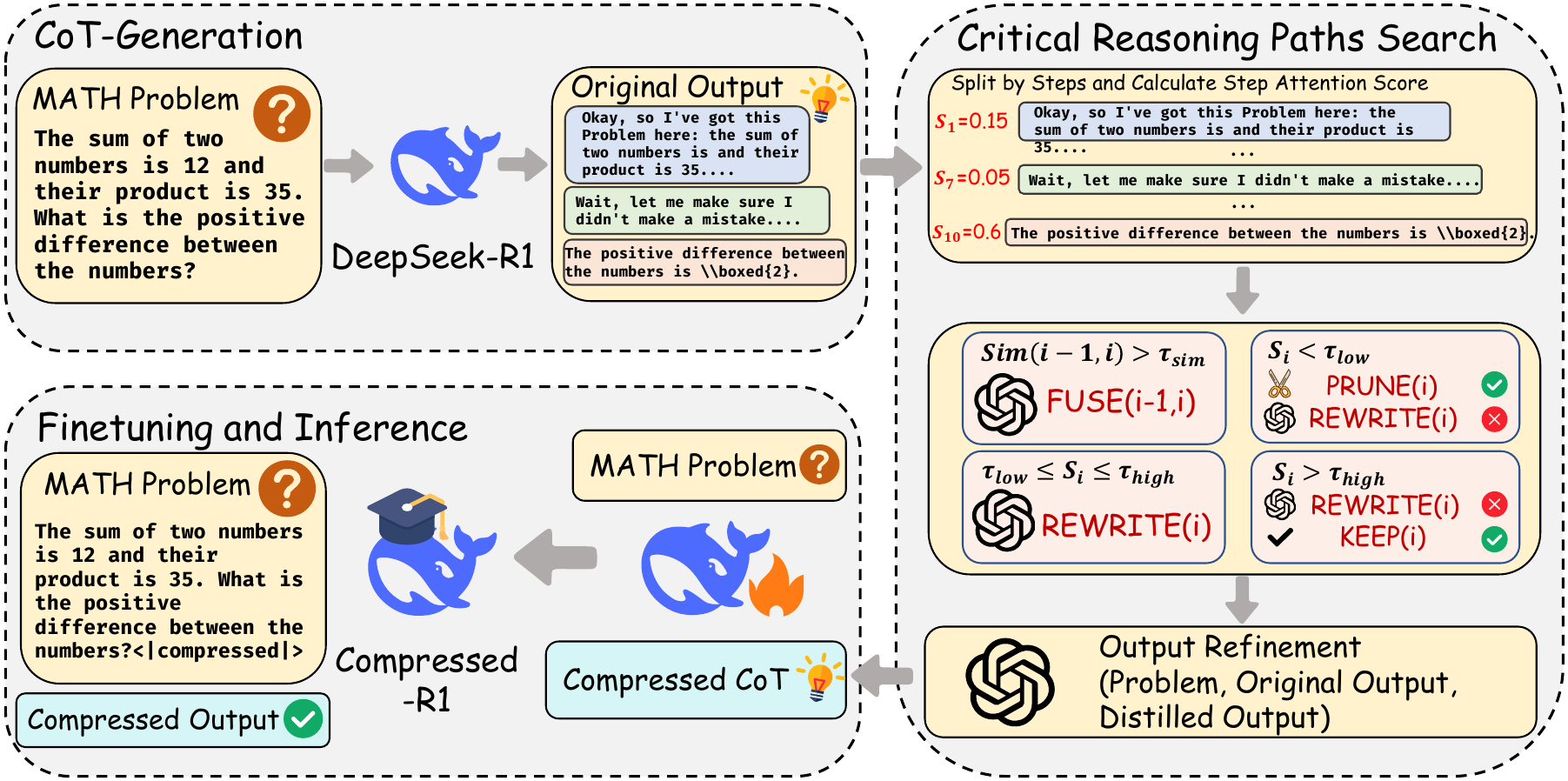} 
    \caption{\textbf{Overview of the CRISP framework.} The process comprises: \textbf{(1) CoT-Generation}, eliciting raw reasoning trajectories from the source model; \textbf{(2) Critical Reasoning Paths Search}, which evaluates step salience via attention scores and distills chains using dynamic operators (\textsc{Keep}, \textsc{Fuse}, \textsc{Prune}, \textsc{Rewrite}) followed by generative refinement; and \textbf{(3) Finetuning and Inference}, employing these refined, high-density trajectories as supervision for efficient reasoning generation.}
    \label{fig:CRISP_illustrations}
     \vspace{-8pt}
\end{figure*}

\section{CRISP: Compressing Redundancy via Intrinsic Saliency Pruning}
Building upon the identification of the \texttt{</think>} token as an informational anchor, we propose \textbf{CRISP} (Figure~\ref{fig:CRISP_illustrations}), a framework designed to distill efficient reasoning paths from the model's internal attention dynamics.

Unlike prevailing compression strategies often rely on external language models for step evaluation, introducing the value misalignment highlighted in Section~\ref{sec:related_works}. In contrast, \textbf{CRISP} shifts from extrinsic supervision to endogenous self-selection. We formulate CoT compression as a structured search governed by intrinsic attention dynamics. By utilizing four atomic operators (\textsc{Keep}, \textsc{Prune}, \textsc{Rewrite}, and \textsc{Fuse}), our framework iteratively constructs a critical reasoning skeleton. Subsequently, an auxiliary model serves as a linguistic refiner to restore syntactic coherence without compromising logical substance. This decoupled design ensures that the compressed CoT retains the target model's inherent reasoning fidelity while achieving textual fluency.

\subsection{Original CoT Generation}
Let $\mathcal{M}_{\theta}$ denote the target Large Language Model parameterized by $\theta$. Given an input query $x$ and a reasoning instruction $\mathcal{I}$, the model generates a response sequence comprising a reasoning chain $\mathcal{R}_{\text{orig}}$ and a subsequent final answer $y$. We formulate this joint generation process as sampling from the model's posterior distribution:
\begin{equation}
(\mathcal{R}_{\text{orig}}, y) \sim P_{\theta}(r, y \mid x, \mathcal{I})
\end{equation}
Here, $\mathcal{R}_{\text{orig}} = \{r_1, r_2, \dots, r_L\}$ consists of $L$ discrete reasoning steps. This sequence encapsulates the comprehensive yet potentially redundant inference process that precipitates the final answer $y$. $\mathcal{R}_{\text{orig}}$ serves as the foundational input for our framework, providing the raw trace from which the essential reasoning signal is distilled.

\subsection{Critical Reasoning Step Search and Compression}
\begin{figure}[t] 
    \centering
    \includegraphics[width=\linewidth]{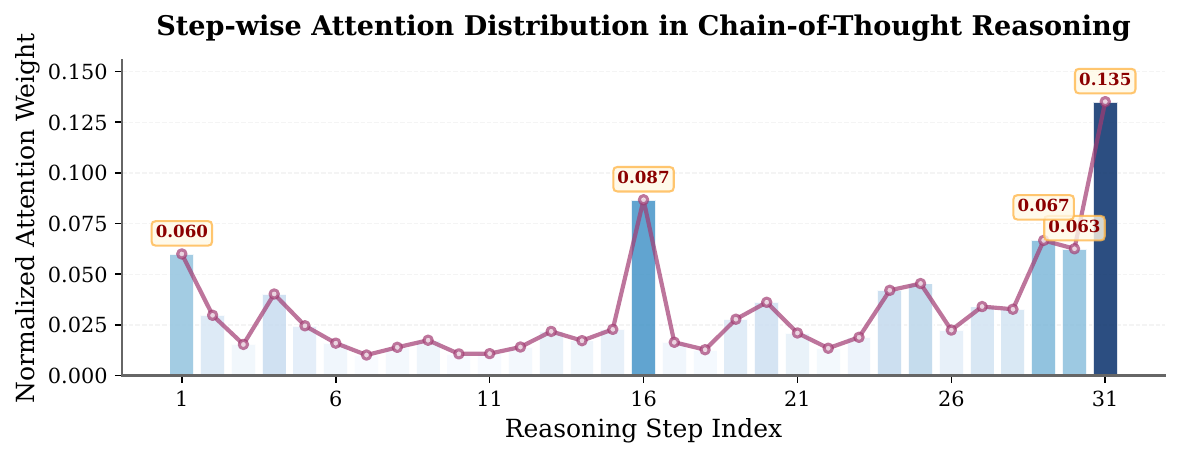}
    \caption{\textbf{Step-wise Attention Distribution in Chain-of-Thought.} The normalized scores $S_i$ exhibit a non-uniform distribution, highlighting that critical information is localized in a few key steps.}
    \label{fig:step_attention_distribution}
    \vspace{-8pt}
\end{figure}

Given the raw reasoning trajectory $\mathcal{R}_{\text{orig}}$ generated by the model, our objective is to distill a compact reasoning skeleton that preserves logical fidelity. We begin by decomposing $\mathcal{R}_{\text{orig}}$ into $L$ discrete steps, $\mathcal{R}_{\text{orig}} = \{r_1, r_2, \dots, r_L\}$, using the standard double newline \verb|\n\n| delimiter. To evaluate the utility of each step without relying on external supervision, we leverage the intrinsic attention dynamics anchored to the \texttt{</think>} token (as detailed in Section \ref{sec:information_anchor}). Since \texttt{</think>} marks the transition from reasoning to the final answer, its attention distribution naturally highlights the antecedent steps most critical to the prediction. Formally, we quantify the endogenous contribution $S_i$ of step $r_i$ by aggregating the attention weights from \texttt{</think>} to all tokens $t_j$ within that step:
\begin{equation}
S_i = \frac{1}{|r_i|} \sum_{t_j \in r_i} \sum_{l=1}^{N_L} \sum_{h=1}^{N_H} \mathbf{A}_{h,l}(\texttt{</think>}, t_j)
\end{equation}
Figure \ref{fig:step_attention_distribution} visualizes the distribution of these attention scores. The plot demonstrates that not all steps hold high salience relative to the input query and the final solution; rather, only a limited subset of the trajectory offers significant contributions to the inference process.
\begin{table}[t]
\centering
\renewcommand{\arraystretch}{1.0} 
\setlength{\tabcolsep}{10pt}      
\begin{tabular}{lc}
\toprule
\textbf{Condition} & \textbf{Allowed Actions ($\mathcal{A}_i$)} \\
\midrule
$\text{Sim}(\mathcal{C}_{\text{last}}, r_i) \ge \tau_{\text{sim}}$ & \textsc{Fuse} \\
$S_i < \tau_{\text{low}}$ & \textsc{Prune}, \textsc{Rewrite} \\
$\tau_{\text{low}} \le S_i \le \tau_{\text{high}}$ & \textsc{Rewrite} \\
$S_i > \tau_{\text{high}}$ & \textsc{Keep}, \textsc{Rewrite} \\
\bottomrule
\end{tabular}
\caption{Mapping of state conditions to the set of allowed actions $\mathcal{A}(\mathcal{S}_i)$. The conditions depend on the score $S_i$ thresholds and semantic similarity.}
\label{tab:allowed_actions}
\vspace{-8pt}
\end{table}
However, relying solely on static threshold-based filtration of $S_i$ is insufficient, as it risks severing critical logical dependencies or retaining redundant, low-density segments. To navigate the trade-off between conciseness and information fidelity, we formulate CoT compression as a structured greedy search. We process the original reasoning trajectory $\mathcal{R}_{\text{orig}} = \{r_1, \dots, r_L\}$ sequentially. At each step $i$, given the partial compressed chain constructed thus far, denoted as $\mathcal{C}$, we select the optimal operator to process the current step $r_i$.

To adaptively mitigate inter-step redundancy and intra-step verbosity, we introduce a dynamic action space $\mathcal{A}_i$. This space comprises non-parametric operators (\textsc{Prune}, \textsc{Keep}), which either discard or preserve the step, and generative operators (\textsc{Rewrite}, \textsc{Fuse}), which leverage an LLM to synthesize concise or merged representations. The admissible actions are constrained by a sequential protocol detailed in Table~\ref{tab:allowed_actions}. Initially, the mechanism assesses semantic redundancy; steps exhibiting high similarity to the context tail $\mathcal{C}_{\text{last}}$ are processed via \textsc{Fuse} to consolidate information. Subsequently, for non-redundant steps, the action space is stratified by the salience score $S_i$: low-salience steps are candidates for \textsc{Prune} or \textsc{Rewrite}, intermediate steps are exclusively refined via \textsc{Rewrite}, while high-salience steps are eligible for either \textsc{Keep} or \textsc{Rewrite}. This structured filtering effectively narrows the search space before reward evaluation.

To select the optimal operator from the allowed set $\mathcal{A}_i$, we employ a reward function that balances the gain in answer likelihood against sequence length. For a candidate action $a$, the reward is calculated as:
\begin{equation}
\begin{split}
R(a) = & \log P_{\theta}(y \mid x, \mathcal{C} \oplus a(r_i)) \\
       & - \log P_{\theta}(y \mid x, \mathcal{C}) - \beta \cdot \text{Len}(a(r_i))
\end{split}
\end{equation}
Here, the first term quantifies the marginal improvement in the model's likelihood of predicting the correct answer $y$ given the action, while the second term imposes a length penalty controlled by the hyperparameter $\beta$. We apply a greedy strategy to iteratively select the action maximizing $R(a)$, appending the result to $\mathcal{C}$ to construct the final reasoning skeleton $R'$. The complete pseudocode for this procedure is detailed in Appendix~\ref{sec:details_of_reasoning_paths_search}.

\subsection{Distilled CoT Refinement}
While the greedy search effectively distills a logical skeleton $R'$, the discrete application of operators can induce syntactic fragmentation or minor logical gaps. To restore textual coherence, we employ a generative refiner $\mathcal{M}_{\text{refine}}$ to reconstruct the final trajectory. The refinement process is formulated as:
\begin{equation}
\mathcal{R}_{\text{CRISP}} = \mathcal{M}_{\text{refine}}(x, R_{orig}, R')    
\end{equation}
Conditioned on the input query $x$ and the original CoT $R_{orig}$, the refiner restores semantic coherence to $\mathcal{R}_{\text{CRISP}}$, smoothing disjointed transitions while strictly preserving the underlying logical substance. This ensures that $\mathcal{R}_{\text{CRISP}}$ attains linguistic fluency without compromising the fidelity of the compressed reasoning path.

\begin{table*}[t]
\centering

\renewcommand{\arraystretch}{1.15}
\setlength{\tabcolsep}{3.5pt}

\resizebox{\textwidth}{!}{%
\begin{tabular}{l ccc ccc ccc ccc}
\toprule
\multirow{2}{*}{\textbf{Method}} & \multicolumn{3}{c}{\textbf{GSM8K}} & \multicolumn{3}{c}{\textbf{MATH-500}} & \multicolumn{3}{c}{\textbf{AMC23}} & \multicolumn{3}{c}{\textbf{Average}} \\
\cmidrule(lr){2-4} \cmidrule(lr){5-7} \cmidrule(lr){8-10} \cmidrule(l){11-13}
& Acc. $\uparrow$ & Tok. $\downarrow$ & TE. $\uparrow$ & Acc. $\uparrow$ & Tok. $\downarrow$ & TE. $\uparrow$ & Acc. $\uparrow$ & Tok. $\downarrow$ & TE. $\uparrow$ & Acc. $\uparrow$ & Tok. $\downarrow$ & TE. $\uparrow$ \\
\midrule

\rowcolor{gray!10} \multicolumn{13}{c}{\textit{\textbf{Model: DeepSeek-R1-Distill-Qwen-1.5B}}} \\
\midrule
Original       & 81.6 & 1669 & 4.89 & 78.2 & 3515 & 2.22 & 60.0 & 5265 & 1.14 & 73.3 & 3483 & 2.10 \\
Truncation     & \underline{73.4} & 1311 & 5.60 & 67.4 & 2629 & 2.56 & 50.0 & 3258 & 1.53 & 63.6 & 2399 & 2.65 \\
CoD            & 70.7 & \underline{677}  & \underline{10.44} & \textbf{78.6} & 2879 & 2.73 & \underline{60.0} & 4462 & 1.34 & \underline{69.8} & 2673 & 2.61 \\
TALE           & 68.7 & 752  & 9.13 & 74.2 & 3257 & 2.28 & \textbf{62.5} & 4812 & 1.19 & 68.5 & 2940 & 2.31 \\
TokenSkip      & 72.9 & 1682 & 4.33 & 66.8 & 3841 & 1.74 & 50.0 & 5202 & 0.96 & 63.2 & 4295 & 1.77 \\
A*-Thought     & 67.9 & 978  & 6.95 & 54.4 & \underline{2015} & \underline{2.70} & 42.5 & \textbf{2528} & \underline{1.68} & 55.0 & 1840 & 2.99 \\
\rowcolor{blue!5} \textbf{CRISP (Ours)} & \textbf{80.6} & \textbf{587} & \textbf{13.73} & \underline{75.0} & \textbf{1813} & \textbf{4.14} & \underline{60.0} & \underline{2607} & \textbf{2.30} & \textbf{71.9} & \textbf{1669} & \textbf{4.31} \\

\midrule \addlinespace[0.5ex] 

\rowcolor{gray!10} \multicolumn{13}{c}{\textit{\textbf{Model: DeepSeek-R1-Distill-Qwen-7B}}} \\
\midrule
Original       & 90.8 & 1376 & 6.60 & 87.4 & 3053 & 2.86 & 72.5 & 4483 & 1.62 & 83.6 & 2971 & 2.81 \\
Truncation     & 84.5 & 1191 & 7.09 & 76.8 & 2419 & 3.17 & 65.0 & 3035 & 2.14 & 75.4 & 2215 & 3.40 \\
CoD            & 71.2 & \underline{279}  & \underline{25.52} & 76.2 & 1841 & \underline{4.14} & \textbf{82.5} & 3217 & \underline{2.56} & \underline{76.6} & 1779 & \underline{4.31} \\
TALE           & 67.9 & \textbf{169}  & \textbf{40.20} & 69.8 & 2254 & 3.10 & \underline{77.5} & 4128 & 1.73 & 71.7 & 2253 & 3.15 \\
TokenSkip      & \underline{84.7} & 1212 & 6.99 & \underline{79.2} & 2636 & 3.00 & 72.5 & 3676 & 1.97 & 74.9 & 2665 & 2.81 \\
A*-Thought     & 75.5 & 663  & 11.39 & 53.0 & \underline{1286} & 4.12 & 37.5 & \textbf{2160} & 1.74 & 55.3 & \underline{1370} & 4.04 \\
\rowcolor{blue!5} \textbf{CRISP (Ours)} & \textbf{90.1} & 374 & 24.09 & \textbf{84.2} & \textbf{1146} & \textbf{7.35} & \underline{77.5} & \underline{2180} & \textbf{3.56} & \textbf{83.9} & \textbf{1235} & \textbf{6.80} \\
\bottomrule
\end{tabular}%
}
\caption{Main results on GSM8K, MATH-500, and AMC23. We report Accuracy (Acc.), Average Tokens (Tok.), and Token Efficiency (TE). \textbf{Bold} and \underline{underline} denote the best and second-best results, respectively.}
\label{tab:main_results}
\vspace{-8pt}
\end{table*}

\subsection{Multi-Task Fine-Tuning and Inference}
Following \citep{yan2025long}, we employ a multi-task fine-tuning strategy governed by a special control token $\kappa$ (denoted as \texttt{<|compressed|>}). To incorporate this control signal, we construct the input $x_{\text{comp}}$ for compressed samples by appending $\kappa$ to the query $x$, wrapped in \texttt{[EOS]} markers:\begin{equation}x_{\text{comp}} = x \oplus \texttt{[EOS]} \kappa \texttt{[EOS]}\end{equation}Let the target sequence $\mathbf{y} = \mathcal{R}_{\text{CRISP}}$ denote the compressed reasoning path. We optimize the model parameters $\theta$ by minimizing the negative log-likelihood over the reasoning tokens:\begin{equation}\mathcal{L} = - \sum_{t=1}^{|\mathbf{y}|} \log P_{\theta}(y_t \mid x_{\text{comp}}, y_{<t})\end{equation}To preserve general reasoning capabilities and mitigate catastrophic forgetting, we mix original trajectories $\mathcal{R}_{\text{orig}}$ into the training corpus, omitting $\kappa$ for these instances. During inference, appending the control sequence $\texttt{[EOS]}\kappa\texttt{[EOS]}$ to the query acts as a steering signal, prompting the model to generate concise, high-density reasoning paths.

\section{Experiments}
\subsection{Experimental Setup}
\paragraph{Models and Training Data.}
To evaluate the efficacy and scalability of CRISP, we conduct experiments on two distilled reasoning models of varying scales: \textbf{DeepSeek-R1-Distill-Qwen-1.5B} and \textbf{DeepSeek-R1-Distill-Qwen-7B}. For the training corpus, we construct a balanced subset from the MATH dataset~\cite{hendrycks2024measuring} via stratified sampling, selecting 500 instances uniformly from each of the five difficulty levels to yield a total of 2,500 samples. We strictly filter these instances to ensure that the ground-truth answer is derivable by the base models and that the maximum sequence length does not exceed 8,192 tokens. Crucially, we leverage the CRISP framework to generate model-specific compression targets rather than employing a unified dataset; this ensures that the supervision signal remains faithful to each model's intrinsic saliency, thereby minimizing distribution shifts.

\paragraph{Evaluation Benchmarks.}
We evaluate our method on three benchmarks spanning varying difficulty levels: \textbf{GSM8K}~\cite{cobbe2021training} for grade-school math, \textbf{MATH-500}~\cite{hendrycks2024measuring} for comprehensive problem solving, and \textbf{AMC23}~\cite{AMC} for competition-level reasoning (see Appendix~\ref{appendix:eval_datasets} for details). Following standard protocols~\cite{guo2025deepseek}, we adopt a consistent decoding strategy with temperature $0.6$ and top-$p$ $0.95$. To accommodate extended reasoning trajectories, we set the maximum generation length to 4,096 tokens for GSM8K and expand it to 8,192 tokens for the more complex MATH-500 and AMC23 datasets. Further implementation details are provided in Appendix~\ref{appendix:inference_and_evaluation}.

\paragraph{Metrics.}
To comprehensively evaluate the trade-off between reasoning capability and computational overhead, we report three metrics: \textbf{Accuracy (Acc.)}, which validates the answer against the ground truth; \textbf{Tokens (Tok.)}, denoting the average length of the generated reasoning trajectories; and \textbf{Token Efficiency (TE.)}. Following~\cite{yan2025long}, Token Efficiency is defined as:
\begin{equation}
    \text{TE} = \frac{\text{Acc}}{\text{Length}} \times 100
\end{equation}
This composite metric captures the accuracy-per-token utility, with higher values indicating more efficient reasoning.
\paragraph{Implementation Details.}
We adopt a unified optimization framework for the proposed model and most baselines, training for 3 epochs with a peak learning rate of $1 \times 10^{-5}$ and a warm-up ratio of 0.1. For the hyperparameters specific to our method, we set the balance coefficient $\beta$ to 0.005 (as in Eq. 4). The attention thresholds $\tau_{high}$ and $\tau_{low}$ are empirically determined based on the top 30\% and bottom 20\% quantiles of the attention distribution, respectively. Additionally, we leverage SimCSE~\cite{gao2021simcse} as the backbone for semantic similarity measurement, setting the similarity threshold $\tau_{sim}$ to 0.7. 
Further details regarding the hardware infrastructure and computational costs are provided in Appendix~\ref{appendix:training_configuration}.

\subsection{Baselines} To validate the effectiveness of CRISP, we benchmark it against a diverse set of redundancy-reduction methods:

\begin{itemize}[leftmargin=*, itemsep=2pt, topsep=2pt, parsep=0pt]
\item \textbf{Truncation}: A baseline method that applies a hard cutoff at a fixed token length, forcing termination strictly based on the token count rather than semantic completion.
\item \textbf{Chain-of-Draft (CoD)}~\cite{xu2025chain}: A prompting strategy mimicking human drafting, which constrains the model to generate information-dense, minimalist reasoning phrases instead of verbose sentences.
\item \textbf{TALE}~\cite{han2025token}: A length-constrained prompting baseline that explicitly instructs the model to reason within a strict token budget via natural language system prompts.
\item \textbf{TokenSkip}~\cite{xia2025tokenskip}: A fine-tuning baseline that employs prompt compression techniques LLMLingua2\cite{pan2024llmlingua} to distill supervision data, enabling the model to learn information-dense reasoning patterns from the pruned CoT trajectories.
\item \textbf{A*-Thought}~\cite{xu2025thought}: A search-based compression framework that treats reasoning compression as a pathfinding problem. It employs an A* search algorithm guided by a bidirectional importance scoring mechanism to identify the optimal subset of reasoning tokens that preserves the deductive chain.
\end{itemize}

\subsection{Results and Analysis}
\paragraph{CRISP achieves a better trade-off between compression and performance.} 
Table \ref{tab:main_results} compares CRISP with the backbone models and representative baselines. Overall, CRISP achieves a favorable balance between reasoning accuracy and inference efficiency. On the 7B benchmark, CRISP achieves an average accuracy of 83.9\%, slightly exceeding the original model's 83.6\%, while reducing token consumption by over 58\% (1235 vs. 2971 tokens). This results in a Token Efficiency (TE) of 6.80, significantly outperforming both the original baseline and other compression methods. In comparison, prompt reduction methods such as TALE and CoD show a notable drop in accuracy, whereas finetuning approaches like TokenSkip yield limited efficiency gains. We also observe consistent performance on the 1.5B scale. Notably, on the AMC23 dataset with the 7B model, CRISP outperforms the original model by a clear margin (77.5\% vs. 72.5\%). This finding suggests that CRISP effectively filters out redundant reasoning steps that may introduce noise in language models, thereby serving as a mechanism for reasoning refinement.

\begin{figure}[t] 
    \centering
    \includegraphics[width=\linewidth]{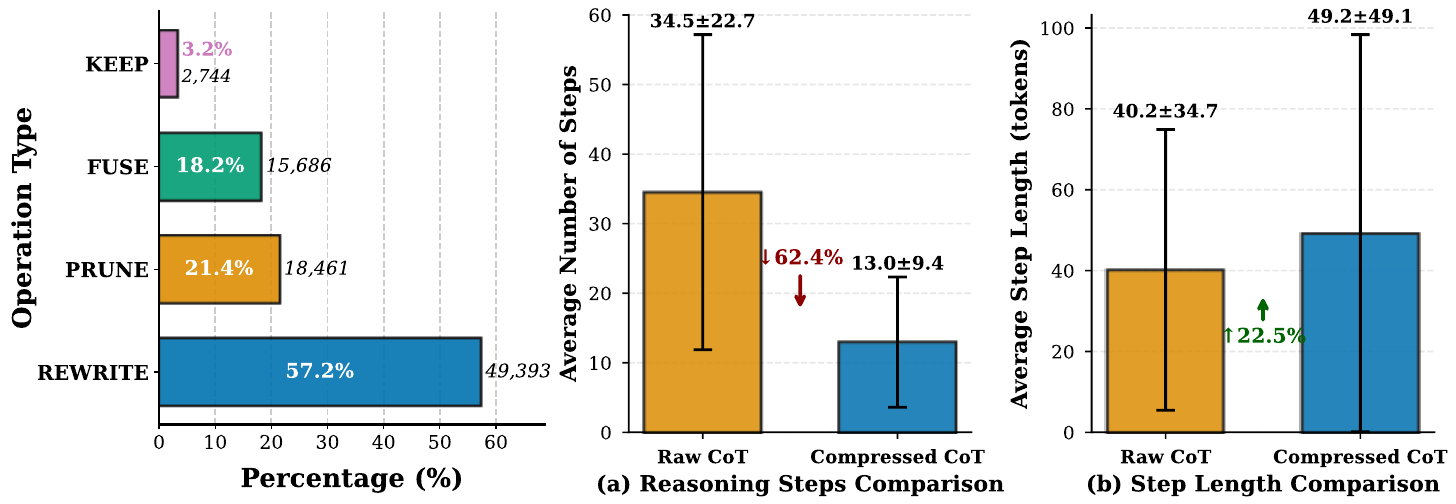}
    \caption{\textbf{Characterization of CRISP compression.} \textbf{Left}: Distribution of atomic operations favoring abstractive synthesis. \textbf{Right}: (a) Reasoning steps decrease by 62.4\%, while (b) average step length increases by 22.5\%, reflecting the consolidation of fragmented logical chains into information-dense units.}
    \label{fig:compression_comparision}
\vspace{-8pt}
\end{figure}

\paragraph{How does CRISP optimize the trade-off between compression and performance?} We dissect the efficiency gains by analyzing the atomic operations in Figure \ref{fig:compression_comparision}. The distribution characterizes an abstractive compression paradigm: \textsc{Rewrite} operations dominate (57.2\%), while \textsc{Keep} actions are negligible (3.2\%). This underscores that CRISP actively synthesizes information to maximize semantic density rather than merely performing extractive pruning. Crucially, this synthesis facilitates a "step-wise consolidation": we observe a sharp reduction in reasoning steps (↓62.4\%) alongside a moderate increase in average step length (↑22.5\%). This suggests that CRISP leverages \textsc{Fuse} (18.2\%) and \textsc{Rewrite} to collapse fragmented logical chains into fewer, high-information reasoning units, streamlining the logical topology while preserving critical evidence.

\begin{table}[t]
\centering
\small
\renewcommand{\arraystretch}{1.1}
\setlength{\tabcolsep}{8pt}

\resizebox{\columnwidth}{!}{%
\begin{tabular}{l ccc}
\toprule
\textbf{Method} & \textbf{Acc.} $\uparrow$ & \textbf{Tok.} $\downarrow$ & \textbf{TE.} $\uparrow$ \\
\midrule

\rowcolor{gray!10} \multicolumn{4}{c}{\textit{\textbf{Model: DeepSeek-R1-Distill-Qwen-1.5B}}} \\
\midrule
Original                    & 78.2 & 3515 & 2.22 \\
A*-Thought                  & 54.4 & \underline{2015} & \underline{2.70} \\
\rowcolor{blue!5} CRISP (w/o Refinement)      & \underline{57.6} & 2265 & 2.54 \\
\rowcolor{blue!5} \textbf{CRISP (Full)} & \textbf{75.0} & \textbf{1813} & \textbf{4.14} \\

\midrule \addlinespace[0.5ex] 

\rowcolor{gray!10} \multicolumn{4}{c}{\textit{\textbf{Model: DeepSeek-R1-Distill-Qwen-7B}}} \\
\midrule
Original                    & 87.4 & 3053 & 2.86 \\
A*-Thought                  & 53.0 & \underline{1286} & 4.12 \\
\rowcolor{blue!5} CRISP (w/o Refinement)      & \underline{70.6} & 1588 & \underline{4.45} \\
\rowcolor{blue!5} \textbf{CRISP (Full)} & \textbf{84.2} & \textbf{1146} & \textbf{7.35} \\

\bottomrule
\end{tabular}%
}
\caption{\textbf{Ablation study on MATH-500.} ``w/o Refinement'' indicates the version with only Reasoning Search. \textbf{Bold} denotes the best results among inference-efficient methods (excluding Original).}
\label{tab:ablation_math500}
\vspace{-10pt}
\end{table}

\paragraph{Ablation Study.} We evaluate the isolated contributions of the core components within CRISP on the MATH-500 benchmark across both 1.5B and 7B parameter scales. As presented in Table \ref{tab:ablation_math500}, the heuristic pruning employed by A*-Thought results in severe performance degradation. In contrast, our reward-guided search preserves the logical backbone more effectively but still suffers from linguistic fragmentation. The Output Refinement module proves critical across both model sizes as it restores semantic coherence. This process recovers accuracy to 75.0\% for the 1.5B model and 84.2\% for the 7B model, retaining over 95\% of the original performance in both cases. Additionally, the refinement stage functions as a secondary compressor by synthesizing disjointed steps. This mechanism reduces token counts to 1813 for the 1.5B model and 1146 for the 7B model, thereby significantly boosting Token Efficiency. These consistent results confirm that coupling topological search with semantic reconstruction is indispensable for maximizing efficiency regardless of model capacity.
\looseness=-1

\paragraph{Impact of CRISP on Reasoning Efficiency.} \label{sec:reasoning_efficiency}
To empirically validate the inference efficiency of CRISP, we analyze the reasoning trajectories of the CRISP-tuned DeepSeek-R1-Distill-Qwen7B compared to the original backbone on MATH-500. By isolating trajectories that lead to correct answers, we decouple efficiency from accuracy. As illustrated in Figure 7 (Left), our method induces a significant distributional shift toward shorter trajectories, demonstrating a tendency for more concise reasoning. Quantitatively, Figure 7 (Right) highlights that to maintain 80\% accuracy, the CRISP-tuned model requires approximately 36 fewer steps than the baseline. This confirms that CRISP effectively mitigates the "overthinking" phenomenon without performance degradation. For a comprehensive analysis covering both 1.5B and 7B model scales across additional datasets, please refer to Appendix~\ref{sec:appendix_trajectory_analysis}.

\begin{figure}[t] 
    \centering
    \includegraphics[width=\linewidth]{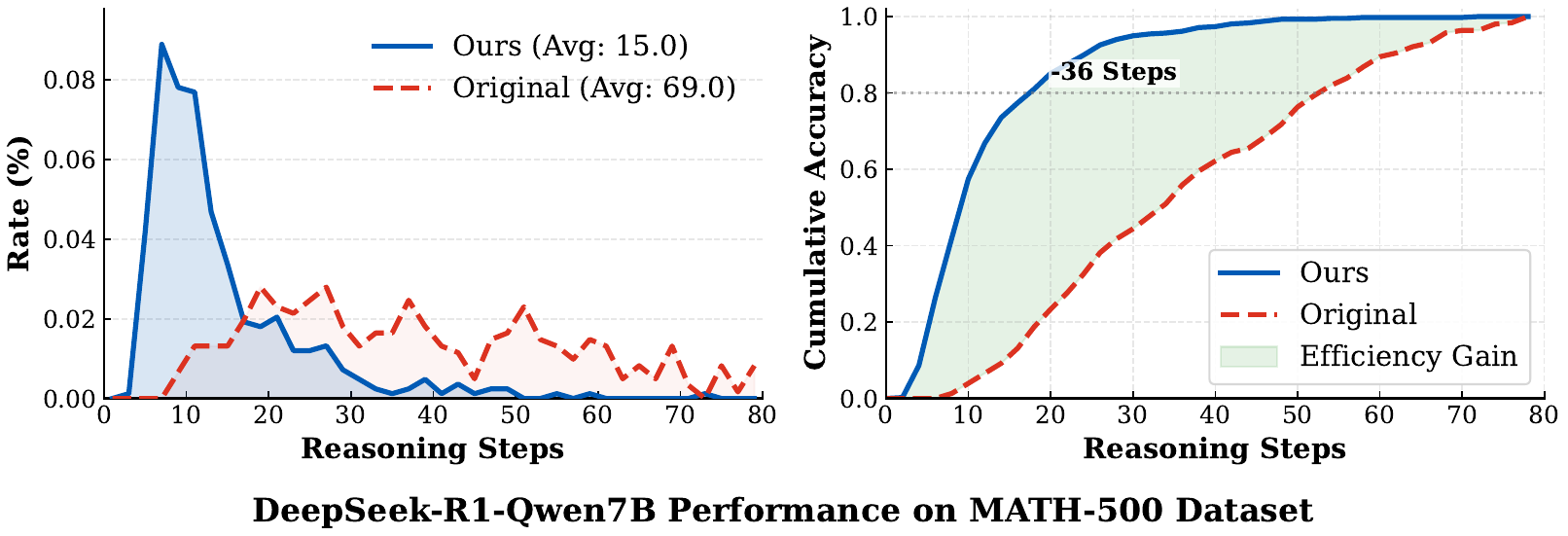}
    \caption{\textbf{Efficiency Analysis on MATH-500.} \textbf{(Left)} Distribution of reasoning steps for correct responses. CRISP significantly compresses the average trajectory length from 69.0 to 15.0. \textbf{(Right)} Cumulative accuracy as a function of steps. The shaded region illustrates the efficiency gain, indicating that CRISP achieves 80\% accuracy using $\sim$36 fewer steps than the baseline.}
    \label{fig:reasoning_distribution}
\vspace{-10pt}
\end{figure}

\section{Conclusion} In this work, we investigate the internal mechanics of R1-series reasoning models, identifying that the reasoning termination token (\texttt{</think>}) functions as a critical information anchor. The attention patterns at this specific position effectively delineate core reasoning trajectories from redundant cognitive computations. Leveraging this insight, we propose \textbf{CRISP}, a framework that steers thought compression via attention-guided search. By optimizing a reward objective that balances answer fidelity with token consumption, and subsequently applying generative refinement for semantic and logical repair, CRISP successfully distills high-density reasoning paths. Extensive experiments demonstrate that models fine-tuned on data synthesized using the CRISP method achieve significant reductions in response length while maintaining robust reasoning performance, offering a scalable solution for efficient LLM deployment.

\section*{Limitations}
Despite the reasoning efficiency demonstrated by CRISP, we acknowledge three primary limitations in our current work. First, our evaluation focuses primarily on mathematical benchmarks characterized by structured logic. The extent to which this reward-guided topological optimization generalizes to open-ended tasks or domains with subjective criteria remains to be investigated. Second, despite the reduced inference cost, the data construction phase incurs significant computational overhead due to iterative search. This offline complexity may present a bottleneck for scaling to extensive datasets. Finally, the effectiveness of the reasoning search is contingent upon the fidelity of reward signals. In specialized or low-resource domains lacking reliable verifiers, the risk of suboptimal path selection could attenuate the quality of the compressed reasoning chains.
\section*{Acknowledgement}
This research is supported by the National Natural Science Foundation of China (No.62476127),  the Natural Science Foundation of Jiangsu Province (No.BK20242039), the Basic Research Program of the Bureau of Science and Technology (ILF24001), the Research Fund (No.PO250624101698), the Scientific Research Starting Foundation of Nanjing University of Aeronautics and Astronautics (No.YQR21022), and the High Performance Computing Platform of Nanjing University of Aeronautics and Astronautics.


\bibliography{custom}

\appendix

\section{Detailed Problem Setup}
\label{sec:Problem Setup}
Given a dataset $\mathcal{D} = \{(x, y)\}$ consisting of input queries $x$ and ground-truth answers $y$, a large language model $\mathcal{M}$ initially generates a verbose reasoning chain $R$ to derive the answer. We denote the length of this reasoning chain as $|R|$.Our objective is to construct a compressed reasoning chain $\hat{R}$ derived from $R$, such that $|\hat{R}| < |R|$. We then fine-tune the model $\mathcal{M}$ on the compressed data $(x, \hat{R}, y)$ to internalize the efficient reasoning pattern. The optimization goal is to find the optimal parameters $\theta$ that minimize the prediction loss on the compressed chains:$$\min_{\theta} \mathbb{E}_{(x, y) \sim \mathcal{D}} \left[ \mathcal{L}(\mathcal{M}_\theta(x, \hat{R}), y) \right]$$where $\mathcal{L}$ denotes the loss function. By solving this optimization problem, we aim to obtain a model that retains the high reasoning performance of the original verbose model while significantly reducing the computational cost in terms of token usage.

\section{More Implementation Details of \textbf{CRISP}}
\label{sec:implementation details}
This appendix details the algorithmic implementation of \textbf{CRISP}. While the complete workflow encompasses the initial CoT generation and the subsequent fine-tuning phase, the core of our methodology lies in the intermediate compression algorithm. This central component operates via a two-stage pipeline. It begins with a greedy search for the critical reasoning path utilizing four atomic primitives (\textsc{Fuse}, \textsc{Prune}, \textsc{Rewrite}, and \textsc{Keep}), followed by a refinement phase that synthesizes the final compressed trajectory. We present the formal algorithmic description and specific implementation nuances below.

\subsection{Details of Critical Reasoning Paths Search and Compression} 
\label{sec:details_of_reasoning_paths_search}

Formally, we model the compression process as a sequential decision-making problem. To avoid the computational intractability of an exhaustive search, we implement a Heuristic Gating Mechanism that dynamically prunes the action space $\mathcal{A}_t$ for each step $r_t$. This mechanism utilizes token-level attention distributions to identify low-contribution steps (triggering \textsc{Prune}/\textsc{Rewrite}) and employs a SimCSE~\cite{gao2021simcse} encoder to detect semantic redundancy (triggering \textsc{Fuse}). The optimal action is then determined by maximizing a joint scoring Reward:
\begin{equation}
\begin{split}
R(a) = & \log P_{\theta}(y \mid x, \mathcal{C} \oplus a(r_i)) \\
       & - \log P_{\theta}(y \mid x, \mathcal{C}) - \beta \cdot \text{Len}(a(r_i))
\end{split}
\end{equation}
which explicitly balances the trade-off between prediction fidelity (log-likelihood gain) and efficiency (token reduction). The complete execution logic is formalized in Algorithm \ref{alg:Critical Reasoning Paths Search}. To ensure reproducibility, we provide the specific instruction templates used for the \textsc{Rewrite} and \textsc{Fuse} operations. These prompts are designed to enforce strict conciseness constraints while preserving logical fidelity. The exact prompt specifications are detailed in Table \ref{tab:prompts}.

\subsection{Details of Distilled CoT Refinement}\label{app:refinement}
While the greedy search algorithm effectively extracts the critical reasoning backbone, the resulting compressed sequence $R'$ frequently exhibits syntactic fragmentation and abrupt logical transitions. Direct training on such disjointed text may compromise the linguistic coherence of the reasoning model. To mitigate this, we introduce a Reference-Conditioned Restoration phase to recover logical fluency while maintaining the structural conciseness of the search result.

Formally, we employ DeepSeek-V3~\cite{liu2024deepseek} to synthesize a refined trajectory $R_{\text{CRISP}}$. We formulate this process as maximizing the conditional probability $P(R_{\text{CRISP}} \mid x, R', R)$, where the generation is conditioned on the joint context of the query $x$, the distilled chain $R'$, and the original chain $R$. This input configuration is critical: conditioning on $R'$ imposes efficiency constraints, while the inclusion of $R$ provides a semantic reference. This dual conditioning enables the model to bridge coherence gaps in $R'$ by reconstructing necessary connective logic and arithmetic details without reverting to the original verbosity. Consequently, the restoration process ensures that $R_{\text{CRISP}}$ maintains high reasoning fidelity with respect to the original logic while strictly adhering to the compression objectives. The specific prompt template used for this restoration process is detailed in Table \ref{tab:refine_prompt_full}.

\subsection{Details of Multi-Task Finetuning and Inference}
To achieve the multi-task fine-tuning objective, we construct training instances in two distinct formats, conditioned on a control signal $\kappa$ (instantiated as \texttt{<|compressed|>}). This dual-task setup allows the model to learn the conditional distribution of compressed reasoning paths while retaining its original reasoning capabilities within a unified parameter space. Table \ref{tab:input_protocols} details the specific input templates used. Notably, for the compression task, the control signal is encapsulated by \texttt{[EOS]} delimiters to establish a clear boundary between the query context and the steering instruction. During inference, the model's generation behavior is modulated by the input structure: appending the control suffix triggers the compressed reasoning mode, whereas providing the raw query prompts the model to generate the original reasoning path.

\begin{table}[h]
\centering
\small
\renewcommand{\arraystretch}{1.5} 
\begin{tabularx}{\columnwidth}{l X}
\toprule
\textbf{Task Type} & \textbf{Input Prompt Construction} \\
\midrule
\multirow{4}{*}{\textbf{Standard}} & \textbf{Input:} \\
& \texttt{\{Question\}} \newline 
  \texttt{Please reason step by step, and put your final answer within \textbackslash boxed\{\}.} \\ 
& $\hookrightarrow$ \textbf{Target:} \texttt{<think> \{\textbf{Original CoT}\} ...} \\
\cmidrule(r){1-1} \cmidrule(l){2-2}
\multirow{4}{*}{\textbf{Compressed}} & \textbf{Input:} \\
& \texttt{\{Question\}} \newline
  \texttt{Please reason step by step, and put your final answer within \textbackslash boxed\{\}}\textbf{\texttt{[EOS]<|compressed|>[EOS]}} \\ 
& $\hookrightarrow$ \textbf{Target:} \texttt{<think> \{\textbf{CRISP CoT}\} ...} \\
\bottomrule
\end{tabularx}
\caption{\textbf{Input formatting protocols for multi-task fine-tuning.} We directly append the specific control suffix to the reasoning prompt for the compressed track, enabling the model to distinguish between the two generation modes.}
\label{tab:input_protocols}
\end{table}

\section{Details about Experiments}

\subsection{Training Datasets}
To construct the training corpus, we curate a balanced subset from the training split of the MATH dataset. Adhering to the generation protocols of \cite{guo2025deepseek}, we synthesize the initial reasoning chains using the base model with temperature $T=0.6$ and sampling top-$p=0.95$, utilizing the standard prompt template defined in Table \ref{tab:input_protocols}. We enforce a correctness constraint by filtering out trajectories that fail to yield the ground-truth answer. From the valid pool, we apply stratified sampling to select 500 instances uniformly from each of the five difficulty levels, resulting in a total of 2,500 samples. These validated sequences constitute our original verbose reasoning chains, denoted as $R_{\text{orig}}$.

\subsection{Evaluation Datasets} 
\label{appendix:eval_datasets}
Our evaluation protocol encompasses three benchmarks representing a broad spectrum of reasoning complexity. We begin with GSM8K~\cite{cobbe2021training}, utilizing the official test set to assess the model's proficiency in basic multi-step arithmetic and logical reasoning. To evaluate performance on more rigorous competition-level mathematics, we employ MATH-500, a representative test subset of the challenging MATH benchmark~\cite{hendrycks2024measuring} covering diverse subjects such as algebra, geometry, and calculus. Finally, to test the model's robustness to out-of-distribution tasks and unseen problems, we introduce AMC23~\cite{AMC}, a dataset comprised of questions from the 2023 American Mathematics Competitions (AMC 10/12). This diverse suite ensures a comprehensive assessment of the model's ability to maintain reasoning fidelity across varying difficulty levels under the compression constraints.

\subsection{Implementation Details}
\label{appendix:implementation_details}
\paragraph{Training Configuration.} 
\label{appendix:training_configuration}
Following the data construction phase, we perform multi-task full-parameter fine-tuning on both the 1.5B and 7B variants of the DeepSeek-R1-Distill series. We utilize the \textbf{LLaMA-Factory} framework~\cite{zheng2024llamafactory} for efficient training. To accommodate memory constraints while maximizing throughput, we employ different DeepSpeed optimization strategies: \textbf{ZeRO-Stage 2} for the 1.5B model and \textbf{ZeRO-Stage 3 (Offload)} for the 7B model~\cite{rasley2020deepspeed}. We maintain a global effective batch size of 32 across all experiments using gradient accumulation. The models are trained for 3 epochs with the AdamW optimizer, a constant learning rate of $1\times 10^{-6}$, and a maximum sequence length of 8,192 tokens. All training runs are conducted in \texttt{bfloat16} precision to ensure numerical stability.

\begin{figure*}[t]
    \centering
    \includegraphics[width=\textwidth]{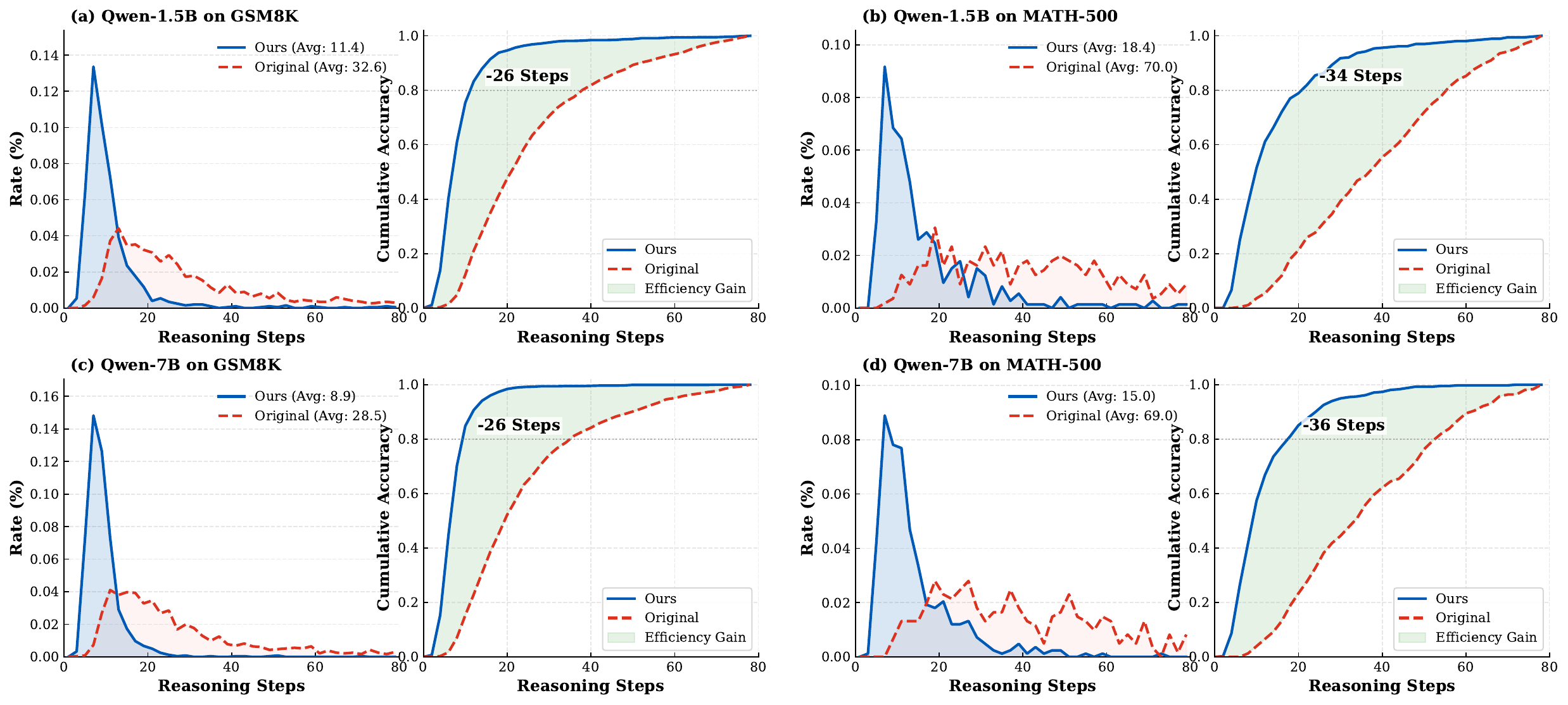}
    \caption{\textbf{Reasoning step distribution and cumulative accuracy analysis across different models and datasets.} 
    (a)-(d) show results for Qwen-1.5B on GSM8K, Qwen-1.5B on MATH-500, Qwen-7B on GSM8K, and Qwen-7B on MATH-500, respectively.
    For each configuration, the left subplot displays the density distribution of reasoning steps for correct samples, 
    while the right subplot shows the cumulative accuracy curves. 
    The figures reveal significant distributional shifts in trajectory length, with the CRISP-tuned models requiring approximately 
    10-15 fewer steps to achieve 80\% cumulative accuracy compared to the original models.}
    \label{fig:reasoning_distribution_all}
\end{figure*}

\paragraph{Inference and Evaluation.} 
\label{appendix:inference_and_evaluation}
For efficient evaluation, we deploy the fine-tuned models using the \textbf{vLLM} engine~\cite{kwon2023efficient}. Following the recommended protocols from \citet{guo2025deepseek}, we adopt a sampling strategy with temperature $T=0.6$ and nucleus sampling top-$p=0.95$. To align with the varying complexity of the evaluation benchmarks, we set the maximum generation length to 4,096 tokens for GSM8K, and extend it to 8,192 tokens for more demanding datasets such as MATH-500, AMC23.

\paragraph{Hardware Infrastructure.}
Our experimental infrastructure comprises two distinct server nodes, equipped with 4 $\times$ NVIDIA L20 and 4 $\times$ NVIDIA RTX 3090 GPUs, respectively. All models are implemented in PyTorch.

\subsection{Implementation Details of Baselines}\label{sec:baseline_details}
To ensure reproducibility and fair comparison, we detail the specific input formulations and training configurations for the baseline methods. Table \ref{tab:baseline_prompts} provides a unified view of the exact input prompts used for CoD~\cite{xu2025chain}, TALE~\cite{han2025token}, A*Thought~\cite{xu2025thought} and TokenSkip~\cite{xia2025tokenskip}. While CoD and TALE rely primarily on prompt engineering to induce brevity, TokenSkip and A*Thought involve specific training procedures, which we detail below.

\textbf{Details for TokenSkip.}
Following the official implementation, we employ LLMLingua2~\cite{pan2024llmlingua} to perform token-level compression on the source data. Crucially, to strictly align the experimental setting, we randomly subsample the compressed data to match the exact training set size of our method. In strict adherence to the original training protocols, we utilize Low-Rank Adaptation ~\cite{hu2022lora} for fine-tuning with a learning rate of $5 \times 10^{-5}$. During inference, we set the compression ratio control parameter $\gamma=0.7$ to guide the generation length.

\textbf{Details for A*Thought.} 
For the A*Thought baseline, we adhere to the original algorithmic search framework. We utilize GPT-2~\cite{radford2019language} as the scorer model and employ the backbone model itself as the validator. Both the training and search phases are conducted using the hyperparameters recommended in the official repository.

\section{Extended Experimental Results and Analysis}
\subsection{Performance Evaluation under Strict Resource Constraints}
While previous experiments have established that the CRISP framework effectively reduces the length of Chain-of-Thought without significantly compromising model capabilities, it is crucial to investigate its viability in strictly resource-constrained scenarios. In this subsection, we focus on CRISP's performance and token efficiency under tighter computational budgets and severe generation limits.

Following the evaluation protocols established by A*Thought~\cite{xu2025thought}, we assess our method on backbone models of varying scales (1.5B and 7B). We simulate low-resource environments by imposing strict maximum token generation limits of 1024, and 2048, respectively. The detailed results across various datasets are presented in \ref{tab:budget_1024} and \ref{tab:budget_2048}.

The empirical results demonstrate CRISP’s effectiveness across different Large Language Models. Notably, CRISP consistently achieves the highest Token Efficiency scores under all budget conditions compared to baselines. This consistent superiority strongly validates the robustness and efficacy of the CRISP method in the CoT compression process, confirming its practical utility even when deployment resources are severely limited.

\subsection{Extended Analysis of Reasoning Trajectory}
\label{sec:appendix_trajectory_analysis}

Complementing the discussion in Section~\ref{sec:reasoning_efficiency}, Figure~\ref{fig:reasoning_distribution_all} presents the comprehensive reasoning step distributions and cumulative accuracy curves across all evaluated configurations (Qwen-1.5B and Qwen-7B on both GSM8K and MATH-500).

These results corroborate the findings presented in the main text, demonstrating that the efficiency gains of CRISP are robust across different model scales and task complexities. As illustrated, the distributional shift towards shorter trajectories is consistent across all settings. Furthermore, the cumulative accuracy analysis confirms that to attain equivalent accuracy levels, the proposed method requires significantly fewer steps. Specifically, a reduction of approximately 25--40 steps is consistently observed across all experimental scenarios.

\subsection{Case Study of CRISP}
To intuitively illustrate the transformation process from the Original CoT to our Compressed CoT, we present a detailed case study in Table \ref{tab:full_case_study}. Drawing from the GSM8K dataset, this example contrasts the raw, full-length CoT (Left) with our processed reasoning path (Right). Specifically, the left column visualizes how the four atomic operations (FUSE, PRUNE, REWRITE, KEEP) are employed to identify the key reasoning path. This is followed by a final optimization stage where an external LLM refines the semantic coherence and logical integrity of the condensed chain.

\subsection{Comprehensive Visualization and Analysis of Attention Dynamics}
\label{sec:appendix_attention}
\begin{figure*}[t]
    \centering
    \includegraphics[width=\textwidth]{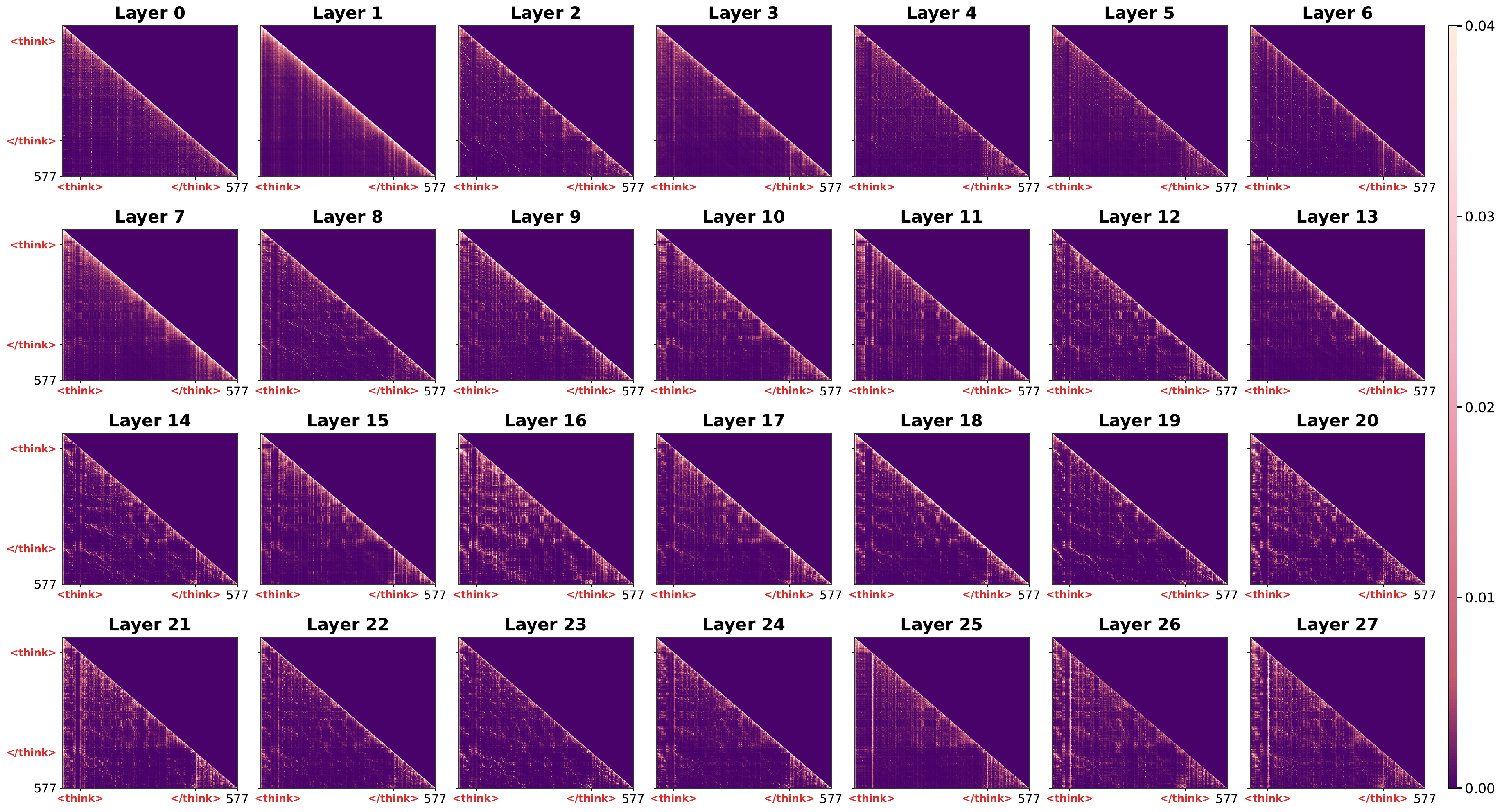}
    \caption{Full layer-wise attention maps for \texttt{DeepSeek-R1-Distill-Qwen-1.5B} on a representative GSM8K sample. Each subplot represents one layer (averaged across all heads). The prominent vertical column emerging in the middle and deep layers corresponds to the \texttt{</think>} token, acting as a site for information aggregation.}
    \label{fig:app_attention_1.5b}
\end{figure*}

\begin{figure*}[t]
    \centering
    \includegraphics[width=\textwidth]{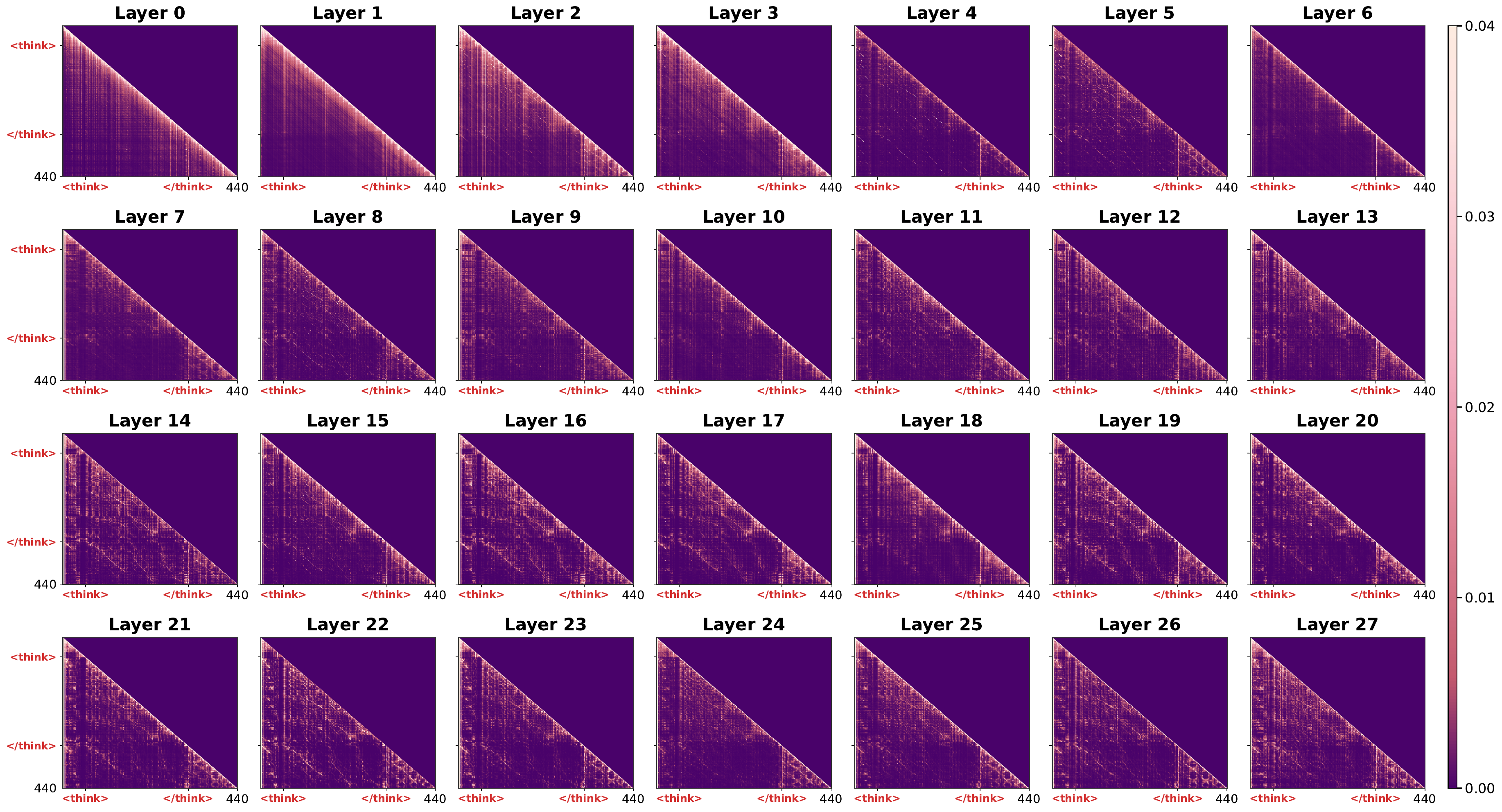}
    \caption{Full layer-wise attention maps for \texttt{DeepSeek-R1-Distill-Qwen-7B}. Similar to the 1.5B variant, the model exhibits a systematic transition from diffuse attention to concentrated attention at the \texttt{</think>} token position as depth increases, reinforcing its role as a semantic anchor for the preceding reasoning chain.}
    \label{fig:app_attention_7b}
\end{figure*}

To provide a comprehensive view of the model's internal mechanics, this section extends the analysis presented in the main text by visualizing the complete layer-wise attention maps for both the \texttt{DeepSeek-R1-Distill-Qwen-1.5B} (Figure \ref{fig:app_attention_1.5b}) and the \texttt{7B} variant (Figure \ref{fig:app_attention_7b}). These visualizations are obtained by averaging the attention scores across all heads within each layer, providing a representative view of how information is processed as it passes through the network.

Across both model scales, we observe a consistent transition in attention patterns as the depth increases. In the initial layers, attention is relatively diffuse, with the model attending broadly to the input tokens and the generated reasoning chain. This suggests that the early stages of processing are primarily concerned with local context and the immediate sequence structure.

As the hidden states progress into the middle and deeper layers, a distinct shift occurs: the attention becomes increasingly concentrated on the \texttt{</think>} token. This vertical concentration indicates that the model begins to treat the boundary token as a primary site for information aggregation. Specifically, during the generation of the final answer, the attention scores for the \texttt{</think>} position remain high, while direct attention to the preceding reasoning steps gradually diminishes. This observation holds for both the 1.5B and 7B models, effectively demonstrating that the use of the \texttt{</think>} token as a semantic anchor is a robust behavior across different parameter scales within the distilled R1 series.

\begin{algorithm*}[t] 
\small
\caption{Critical Reasoning Paths Search}
\label{alg:Critical Reasoning Paths Search}
\SetKwInOut{Input}{Input}
\SetKwInOut{Output}{Output}
\SetKwFunction{Update}{Update}
\SetKwFunction{Sim}{Sim}

\Input{Query $x$, Verbose Chain $R=\{r_1, \dots, r_N\}$, Model $\mathcal{M}$, Efficiency weight $\beta$, Thresholds $\tau$}
\Output{Compressed Chain $\mathcal{C}$}

\tcp{1. Pre-computation Phase}
Compute step-wise attention scores $\mathcal{S} = \{s_1, \dots, s_N\}$ for the original chain using $\mathcal{M}$\;
Initialize compressed chain $\mathcal{C} \leftarrow \emptyset$\;

\tcp{2. Sequential Greedy Search}
\For{$t \leftarrow 1$ \KwTo $N$}{
    Let $r_t$ be the current step and $s_t \in \mathcal{S}$ its attention score\;
    
    \tcp{Heuristic Gating: Determine Allowed Action Space $\mathcal{A}_t$}
    \uIf{\textbf{not} IsEmpty($\mathcal{C}$) \textbf{and} $\Sim( \mathcal{C}_{last}, r_t) \ge \tau_{sim}$}{
        $\mathcal{A}_t \leftarrow \{ \textsc{Fuse} \}$ \tcp*{Merge semantically similar steps}
    }
    \uElseIf{$s_t < \tau_{low}$}{
        $\mathcal{A}_t \leftarrow \{ \textsc{Prune}, \textsc{Rewrite} \}$ \tcp*{Skip or Condense low-info steps}
    }
    \uElseIf{$s_t < \tau_{high}$}{
        $\mathcal{A}_t \leftarrow \{ \textsc{Rewrite} \}$ \tcp*{Condense medium-info steps}
    }
    \uElse{
        $\mathcal{A}_t \leftarrow \{ \textsc{Keep}, \textsc{Rewrite} \}$ \tcp*{Preserve critical logic}
    }
    
    \tcp{Action Selection: Joint Optimization of Fidelity and Efficiency}
    $a^* \leftarrow \operatorname*{argmax}_{a \in \mathcal{A}_t} \Big( \underbrace{\Delta \log P(y|x, \mathcal{C} \oplus a(r_t))}_{\text{Prediction Fidelity Gain}} - \beta \cdot \underbrace{\text{Len}(a(r_t))}_{\text{Token Cost}} \Big)$\;
    
    \tcp{State Update}
    $\mathcal{C} \leftarrow \Update(\mathcal{C}, a^*, r_t)$\;
}
\Return{$\mathcal{C}$}
\end{algorithm*}

\begin{table*}[t!]
\centering
\small 
\renewcommand{\arraystretch}{1.5} 
\begin{tabularx}{\textwidth}{l X} 
\toprule
\textbf{Method} & \textbf{Input Prompt Construction} \\
\midrule
\multirow{3}{*}{\textbf{Standard / A*Thought}} & \textbf{Input:} \\
 & \texttt{\{Question\}} \newline
   \texttt{Please reason step by step, and put your final answer within \textbackslash boxed\{\}.} \\
\midrule
\multirow{5}{*}{\textbf{CoD}~\cite{xu2025thought}} & \textbf{Input:} \\
 & \texttt{\{Question\}} \newline
   \texttt{Please reason step by step, and put your final answer within \textbackslash boxed\{\}.} \newline
   \texttt{Think step by step, but only keep a minimum draft for each thinking step, with 5 words at most.} \\
\midrule
\multirow{3}{*}{\textbf{TALE}~\cite{han2025token}} & \textbf{Input:} \\
 & \texttt{\{Question\}} \newline
   \texttt{Let's think step by step and use less than 4096 tokens.} \\
\midrule
\multirow{4}{*}{\textbf{TokenSkip}~\cite{xia2025tokenskip}} & \textbf{Input:} \\
 & \texttt{\{Question\}} \newline
   \texttt{Please reason step by step, and put your final answer within \textbackslash boxed\{\}}\texttt{[EOS]\{Ratio\}[EOS]} \\
\midrule
\multirow{4}{*}{\textbf{CRISP (Ours)}} & \textbf{Input:} \\
 & \texttt{\{Question\}} \newline
   \texttt{Please reason step by step, and put your final answer within \textbackslash boxed\{\}}\textbf{\texttt{[EOS]<|compressed|>[EOS]}} \\
\bottomrule
\end{tabularx}
\caption{The exact input prompt templates used for the baselines and our method. Note that \textbf{TALE} replaces the standard instruction with a length-constrained prompt, whereas \textbf{CoD}, \textbf{TokenSkip}, and \textbf{CRISP} append specific constraints or control tokens to the standard instruction. The control suffix for our method is highlighted in bold.}
\label{tab:baseline_prompts}
\end{table*}

\begin{table*}[t]
    \centering
    \small
    \renewcommand{\arraystretch}{1.3} 
    \begin{tabularx}{\textwidth}{p{0.12\textwidth} X}
        \toprule
        \textbf{Primitive} & \textbf{Prompt Template} \\
        \midrule
        
        \multirow{8}{*}{\textsc{Rewrite}} 
        & \textbf{[System Message]} \\
        & You are an expert at condensing reasoning steps. Your task is to rewrite the given reasoning step to be more concise while preserving all essential information and logical flow. \\
        & \textit{Rules:} \\
        & 1. Keep all key facts, numbers, and logical connections. \\
        & 2. Remove redundant phrases and verbose expressions. \\
        & 3. Maintain the mathematical or logical correctness. \\
        & 4. Output ONLY the condensed step, no explanations. \\
        \cmidrule(l){2-2} 
        & \textbf{[User Message]} \\
        & Compress this reasoning step as short as possible: \\
        & \texttt{<step> \{Original Step\} </step>} \\
        & Compressed: \\
        \midrule
        
        \multirow{9}{*}{\textsc{Fuse}} 
        & \textbf{[System Message]} \\
        & You are an expert at merging reasoning steps. Your task is to combine two consecutive reasoning steps into a single, coherent step while preserving all essential information. \\
        & \textit{Rules:} \\
        & 1. Preserve all key facts, numbers, and calculations. \\
        & 2. Maintain logical flow and correctness. \\
        & 3. Remove redundant information that appears in both steps. \\
        & 4. The merged step should be shorter than the sum of both steps. \\
        & 5. Output ONLY the merged step, no explanations. \\
        \cmidrule(l){2-2} 
        & \textbf{[User Message]} \\
        & Merge these two steps into one step as short as possible: \\
        & Step 1: \texttt{\{Step 1\}} \\
        & Step 2: \texttt{\{Step 2\}} \\
        & Merged: \\
        
        \bottomrule
    \end{tabularx}
    \caption{\textbf{Prompt templates used for Atomic Operations.} We utilize structured system instructions to enforce strict constraints on information retention and output brevity. The \texttt{\{...\}} placeholders denote the dynamic input content from the reasoning chain.}
    \label{tab:prompts}
\end{table*}

\begin{table*}[t!]
\centering
\small
\renewcommand{\arraystretch}{1.25} 
\begin{tabularx}{\textwidth}{l X}
\toprule
\textbf{Role} & \textbf{Content Specification} \\
\midrule
\textbf{System} & You are an expert mathematical editor. Your task is to refine a rough reasoning draft. Restore logical continuity and mathematical accuracy. Match the Original CoT's exact tone, formatting, and style. \\
\midrule
\textbf{User} & \textbf{\#\#\# Question} \newline
\texttt{\{Input Query\}} \vspace{0.5em} \newline
\textbf{\#\#\# Original CoT (ONLY for Reference)} \newline
\texttt{\{Original Verbose Chain\}} \vspace{0.5em} \newline
\textbf{\#\#\# Rough Draft (To Refine)} \newline
\texttt{\{Compressed Output from Search\}} \vspace{0.5em} \newline
\textbf{\#\#\# Instruction} \newline
Refine the Rough Draft to ensure mathematical coherence and logical flow. \newline
1. Fill in missing algebraic manipulations and arithmetic calculations. \newline
2. Match the style and formatting of the Original CoT. \newline
3. Output ONLY the refined reasoning text. \newline
4. Ensure the calculations lead correctly to the final answer. \vspace{0.5em} \newline
\textbf{\#\#\# Refined Rough Solution:} \\
\bottomrule
\end{tabularx}
\caption{The full prompt template used in the \textbf{Reference-Conditioned Restoration} phase. The prompt is structured to condition the teacher model on three inputs: the query, the semantic reference (Original CoT), and the structural target (Distilled CoT), ensuring the output is both fluent and concise.}
\label{tab:refine_prompt_full}
\end{table*}

\begin{table*}[t]
\centering
\renewcommand{\arraystretch}{1.15}
\setlength{\tabcolsep}{3.5pt}

\resizebox{\textwidth}{!}{%
\begin{tabular}{l ccc ccc ccc ccc}
\toprule
\multirow{2}{*}{\textbf{Method}} & \multicolumn{3}{c}{\textbf{GSM8K}} & \multicolumn{3}{c}{\textbf{MATH-500}} & \multicolumn{3}{c}{\textbf{AMC23}} & \multicolumn{3}{c}{\textbf{Average}} \\
\cmidrule(lr){2-4} \cmidrule(lr){5-7} \cmidrule(lr){8-10} \cmidrule(l){11-13}
& Acc. $\uparrow$ & Tok. $\downarrow$ & TE. $\uparrow$ & Acc. $\uparrow$ & Tok. $\downarrow$ & TE. $\uparrow$ & Acc. $\uparrow$ & Tok. $\downarrow$ & TE. $\uparrow$ & Acc. $\uparrow$ & Tok. $\downarrow$ & TE. $\uparrow$ \\
\midrule

\rowcolor{gray!10} \multicolumn{13}{c}{\textit{\textbf{Model: DeepSeek-R1-Distill-Qwen-1.5B (Budget: 1024)}}} \\
\midrule
Original    & 53.4 & 910 & 5.86 & 22.0 & 1007 & 2.18 & 10.0 & 1023 & 0.98 & 28.5 & 980 & 2.91 \\
CoD         & \underline{67.5} & 472 & \underline{14.30} & \underline{36.8} & 926 & 3.97 & 12.5 & 997 & 1.25 & 38.9 & 798 & 4.87 \\
TALE        & 56.3 & \underline{454} & 12.41 & 22.8 & 944 & 2.42 & 17.5 & 995 & 1.76 & 32.2 & 798 & 4.04 \\
TokenSkip   & 54.6 & 822 & 6.64 & 29.6 & 958 & 3.09 & 10.0 & 1003 & 1.00 & 31.4 & 928 & 3.38 \\
A*-Thought  & 63.1 & 638 & 9.89 & 36.2 & \textbf{758} & \underline{4.78} & \underline{15.0} & \underline{960} & \underline{1.56} & \underline{38.1} & \underline{785} & \underline{4.85} \\
\rowcolor{blue!5} \textbf{CRISP (Ours)} & \textbf{78.1} & \textbf{431} & \textbf{18.12} & \textbf{55.8} & \underline{765} & \textbf{7.29} & \textbf{52.5} & \textbf{927} & \textbf{5.66} & \textbf{62.1} & \textbf{708} & \textbf{8.77} \\

\midrule \addlinespace[0.5ex] 

\rowcolor{gray!10} \multicolumn{13}{c}{\textit{\textbf{Model: DeepSeek-R1-Distill-Qwen-7B (Budget: 1024)}}} \\
\midrule
Original    & 58.4 & 897 & 6.51 & 22.8 & 1002 & 2.28 & 17.5 & 1020 & 1.72 & 32.9 & 973 & 3.38 \\
CoD         & \underline{72.8} & \underline{262} & \underline{27.78} & \underline{46.6} & 775 & 6.01 & 22.5 & 925 & 2.43 & 47.3 & 654 & \underline{7.23} \\
TALE        & 70.7 & \textbf{160} & \textbf{44.21} & 34.6 & \underline{684} & 5.06 & 12.5 & 962 & 1.30 & 39.3 & \textbf{602} & 6.53 \\
TokenSkip   & 71.5 & 747 & 9.57 & 38.4 & 933 & 4.12 & \underline{32.5} & 957 & \underline{3.40} & \underline{47.5} & 879 & 5.40 \\
A*-Thought  & 71.6 & 564 & 12.69 & 41.6 & \textbf{653} & \underline{6.37} & 17.5 & \textbf{694} & 2.52 & 43.6 & \underline{637} & 6.84 \\
\rowcolor{blue!5} \textbf{CRISP (Ours)} & \textbf{89.2} & 369 & 24.18 & \textbf{66.2} & 715 & \textbf{9.26} & \textbf{42.5} & \underline{873} & \textbf{4.87} & \textbf{66.0} & 652 & \textbf{10.12} \\

\bottomrule
\end{tabular}%
}
\caption{Performance comparison under a strict token budget of \textbf{1024 tokens}. We report Accuracy (Acc.), Average Tokens (Tok.), and Token Efficiency (TE). \textbf{Bold} and \underline{underline} denote the best and second-best results, respectively.}
\label{tab:budget_1024}
\end{table*}

\begin{table*}[t]
\centering
\renewcommand{\arraystretch}{1.15}
\setlength{\tabcolsep}{3.5pt}

\resizebox{\textwidth}{!}{%
\begin{tabular}{l ccc ccc ccc ccc}
\toprule
\multirow{2}{*}{\textbf{Method}} & \multicolumn{3}{c}{\textbf{GSM8K}} & \multicolumn{3}{c}{\textbf{MATH-500}} & \multicolumn{3}{c}{\textbf{AMC23}} & \multicolumn{3}{c}{\textbf{Average}} \\
\cmidrule(lr){2-4} \cmidrule(lr){5-7} \cmidrule(lr){8-10} \cmidrule(l){11-13}
& Acc. $\uparrow$ & Tok. $\downarrow$ & TE. $\uparrow$ & Acc. $\uparrow$ & Tok. $\downarrow$ & TE. $\uparrow$ & Acc. $\uparrow$ & Tok. $\downarrow$ & TE. $\uparrow$ & Acc. $\uparrow$ & Tok. $\downarrow$ & TE. $\uparrow$ \\
\midrule

\rowcolor{gray!10} \multicolumn{13}{c}{\textit{\textbf{Model: DeepSeek-R1-Distill-Qwen-1.5B (Budget: 2048)}}} \\
\midrule
Original    & 71.2 & 1310 & 5.43 & 45.4 & 1754 & 2.59 & 32.5 & 1913 & 1.70 & 49.7 & 1659 & 3.00 \\
CoD         & \underline{71.3} & \underline{576} & \underline{12.39} & \underline{59.2} & 1468 & 4.03 & \underline{35.0} & 1812 & \underline{1.93} & \underline{55.2} & 1285 & 4.30 \\
TALE        & 64.3 & 645 & 9.97 & 47.4 & 1599 & 2.96 & 27.5 & 1806 & 1.52 & 46.4 & 1350 & 3.44 \\
TokenSkip   & 67.4 & 1206 & 5.59 & 48.6 & 1633 & 2.98 & 27.5 & 1756 & 1.57 & 47.8 & 1532 & 3.12 \\
A*-Thought  & 66.5 & 779 & 8.54 & 48.0 & \underline{1102} & \underline{4.36} & 22.5 & \textbf{1309} & 1.72 & 45.7 & \underline{1063} & \underline{4.30} \\
\rowcolor{blue!5} \textbf{CRISP (Ours)} & \textbf{79.5} & \textbf{482} & \textbf{16.50} & \textbf{68.0} & \textbf{1057} & \textbf{6.43} & \textbf{50.0} & \underline{1311} & \textbf{3.81} & \textbf{65.8} & \textbf{950} & \textbf{6.93} \\

\midrule \addlinespace[0.5ex] 

\rowcolor{gray!10} \multicolumn{13}{c}{\textit{\textbf{Model: DeepSeek-R1-Distill-Qwen-7B (Budget: 2048)}}} \\
\midrule
Original    & 83.6 & 1178 & 7.09 & 52.8 & 1713 & 3.08 & 40.0 & 1854 & 2.16 & 58.8 & 1582 & 3.72 \\
CoD         & 72.0 & \underline{275} & \underline{26.16} & \underline{61.2} & 1087 & 5.63 & \underline{47.5} & 1575 & \underline{3.02} & 60.2 & 979 & \underline{6.15} \\
TALE        & 69.0 & \textbf{164} & \textbf{42.07} & 49.8 & 1099 & 4.53 & 42.5 & 1654 & 2.57 & 53.8 & 972 & 5.53 \\
TokenSkip   & \underline{82.9} & 942 & 8.80 & 60.4 & 1437 & 4.20 & \underline{47.5} & 1615 & 2.94 & \underline{63.6} & 1331 & 4.78 \\
A*-Thought  & 74.8 & 620 & 12.06 & 48.4 & \textbf{804} & \underline{6.02} & 25.0 & \textbf{1040} & 2.40 & 49.4 & \textbf{821} & 6.02 \\
\rowcolor{blue!5} \textbf{CRISP (Ours)} & \textbf{89.9} & 369 & 24.37 & \textbf{78.0} & \underline{895} & \textbf{8.72} & \textbf{65.0} & \underline{1275} & \textbf{5.10} & \textbf{77.6} & \underline{846} & \textbf{9.17} \\

\bottomrule
\end{tabular}%
}
\caption{Performance comparison under a token budget of \textbf{2048 tokens}. We report Accuracy (Acc.), Average Tokens (Tok.), and Token Efficiency (TE). \textbf{Bold} and \underline{underline} denote the best and second-best results, respectively.}
\label{tab:budget_2048}
\end{table*}

\begin{table*}[t]
    \centering
    \scriptsize 
    \renewcommand{\arraystretch}{1.2}
    \setlength{\tabcolsep}{6pt}
    \begin{tabular}{p{0.48\textwidth} | p{0.48\textwidth}}
        \toprule
        \textbf{Original CoT (Annotated with Search Actions)} & \textbf{Evolution of Reasoning (Compressed $\to$ Refined)} \\
        \midrule
        
        \textbf{Question:} What is the positive difference between the sum of $\frac{1}{2}$ and $\frac{1}{3}$ and the product of $\frac{1}{2}$ and $\frac{1}{3}$? \newline
        \vspace{0.2em}
        
        \stepRewrite{Alright, so I need to find the positive difference between the sum of 1/2 and 1/3 and the product of 1/2 and 1/3. Hmm, okay, let's break this down step by step. I think I should first figure out what the sum is and then the product, and then subtract one from the other. But wait, the question says the positive difference, so maybe I need to make sure the result is positive, just in case the subtraction goes the other way. Let me write this down to visualize it better.} \par \vspace{0.3em}
        
        \stepRewrite{First, let's calculate the sum of 1/2 and 1/3. I remember that to add fractions, they need to have the same denominator. The denominators here are 2 and 3. The least common denominator for 2 and 3 is 6. So I can convert both fractions to have 6 as the denominator.} \par \vspace{0.3em}
        
        \stepFuse{1/2 is equal to 3/6 because 1 times 3 is 3 and 2 times 3 is 6. Similarly, 1/3 is equal to 2/6 because 1 times 2 is 2 and 3 times 2 is 6. Now, adding these together: 3/6 + 2/6. Since the denominators are the same, I can just add the numerators. 3 plus 2 is 5, so the sum is 5/6. Got that part down.} \par \vspace{0.3em}
        
        \stepFuse{Next, I need to find the product of 1/2 and 1/3. Multiplying fractions is straightforward, right? Multiply the numerators together and the denominators together. So, 1 times 1 is 1, and 2 times 3 is 6. Therefore, the product is 1/6. That seems simple enough.} \par \vspace{0.3em}
        
        \stepRewrite{Now, the problem asks for the positive difference between the sum and the product. So, I think that means I need to subtract the product from the sum and then take the positive value of that result. Let me write that down: positive difference = |sum - product|.} \par \vspace{0.3em}
        
        \stepPrune{So, plugging in the values I found: sum is 5/6 and product is 1/6. Therefore, the difference is 5/6 - 1/6. Again, since the denominators are the same, I can subtract the numerators directly. 5 minus 1 is 4, so the difference is 4/6.} \par \vspace{0.3em}
        
        \stepPrune{Wait, 4/6 can be simplified, right? Both 4 and 6 are divisible by 2. So, dividing numerator and denominator by 2, we get 2/3. Therefore, the positive difference is 2/3.} \par \vspace{0.3em}
        
        \stepRewrite{Let me just double-check my steps to make sure I didn't make a mistake. First, adding 1/2 and 1/3: converted to sixths, got 3/6 and 2/6, added to 5/6. Then, multiplying 1/2 and 1/3: 1/6. Then subtracting 1/6 from 5/6 gives 4/6, which simplifies to 2/3. Seems solid.} \par \vspace{0.3em}
        
        \stepRewrite{Hmm, another way to think about it: maybe I should write it out as an equation. Let me define S as the sum and P as the product.} \par \vspace{0.3em}
        
        \stepRewrite{S = 1/2 + 1/3 \newline P = (1/2) * (1/3) \newline Difference = |S - P|} \par \vspace{0.3em}
        
        \stepRewrite{Calculating S: \newline 1/2 + 1/3 = (3/6) + (2/6) = 5/6} \par \vspace{0.3em}
        
        \stepFuse{Calculating P: \newline (1/2)*(1/3) = 1/6} \par \vspace{0.3em}
        
        \stepRewrite{Difference: \newline |5/6 - 1/6| = |4/6| = 4/6 = 2/3} \par \vspace{0.3em}
        
        \stepRewrite{Yep, that's consistent with what I did before. So, I think 2/3 is the correct answer. Just to make sure, let me think about the operations again. Sum is bigger than the product because both fractions are less than 1, so their product will be smaller than either of them. So, subtracting the product from the sum should give a positive number, which it does, 2/3. So, that makes sense.} \par \vspace{0.3em}
        
        \stepFuse{Alternatively, if I had subtracted the sum from the product, I would have gotten a negative number, but since the question asks for the positive difference, it's the absolute value, so it's 2/3 either way. But in this case, since sum is bigger, it's just 2/3.} \par \vspace{0.3em}
        
        \stepRewrite{I don't think I made any calculation errors here. The steps all follow logically, and I've double-checked them. So, I'm confident that the positive difference is 2/3.} \par \vspace{0.3em}
        
        \stepKeep{**Final Answer** \newline The positive difference is $\boxed{\dfrac{2}{3}}$.}
        
        & 
        \textbf{1. Compressed Output (Post-Search)} \newline
        \textit{(Result of executing search actions)} \newline
        \vspace{0.5em}
        
        Find the positive difference between (1/2 + 1/3) and (1/2 $\times$ 1/3).
        
        To add 1/2 and 1/3, find a common denominator of 6, converting them to 3/6 and 2/6, then add the numerators to get 5/6; then find their product by multiplying the numerators (1$\times$1=1) and denominators (2$\times$3=6) to get 1/6.
        
        Positive difference = |sum - product|.
        
        Adding 1/2 and 1/3 gives 5/6. Multiplying them gives 1/6. Subtracting 1/6 from 5/6 yields 4/6, which simplifies to 2/3.
        
        Define S as the sum and P as the product.
        
        S = 1/2 + 1/3, P = (1/2)*(1/3), Difference = |S - P|
        
        S = 1/2 + 1/3 = 3/6 + 2/6 = 5/6, and P = (1/2)*(1/3) = 1/6.
        
        |5/6 - 1/6| = 4/6 = 2/3
        
        Since the sum exceeds the product, the positive difference is 2/3.
        
        The positive difference is 2/3.
        
        **Final Answer**
        The positive difference is $\boxed{\dfrac{2}{3}}$.
        
        \vspace{1.5em}
        \hrule
        \vspace{0.2em}
        \centering $\downarrow$ \textbf{Refinement Step} $\downarrow$ 
        \vspace{0.2em}
        \hrule
        \vspace{1.5em}
        
        \raggedright
        \textbf{2. Final Refined CoT (Training Target)} \newline
        Alright, so I need to find the positive difference between the sum of $\frac{1}{2}$ and $\frac{1}{3}$ and the product of $\frac{1}{2}$ and $\frac{1}{3}$.
        Let’s break this down step by step.
        
        First, I’ll compute the sum:
        \[ \frac{1}{2} + \frac{1}{3}. \]
        To add these fractions, I need a common denominator. The least common multiple of 2 and 3 is 6.
        \[ \frac{1}{2} = \frac{1 \times 3}{2 \times 3} = \frac{3}{6}, \quad \frac{1}{3} = \frac{1 \times 2}{3 \times 2} = \frac{2}{6}. \]
        Adding them:
        \[ \frac{3}{6} + \frac{2}{6} = \frac{3+2}{6} = \frac{5}{6}. \]
        
        Next, I’ll compute the product:
        \[ \frac{1}{2} \times \frac{1}{3} = \frac{1 \times 1}{2 \times 3} = \frac{1}{6}. \]
        
        Now, the problem asks for the positive difference between the sum and the product. That means I need
        \[ \left| \text{sum} - \text{product} \right|. \]
        Substituting the values:
        \[ \left| \frac{5}{6} - \frac{1}{6} \right| = \left| \frac{4}{6} \right| = \frac{4}{6}. \]
        This fraction can be simplified by dividing numerator and denominator by 2:
        \[ \frac{4}{6} = \frac{2}{3}. \]
        
        Let me double-check:
        Since $\frac{5}{6} > \frac{1}{6}$, the difference is already positive, so the absolute value doesn’t change the result.
        Thus, the positive difference is $\frac{2}{3}$.
        
        \textbf{Final Answer}
        \[ \boxed{\dfrac{2}{3}} \]
        \\
        \tabularnewline
    \end{tabular}
    \caption{
Case study illustrating the complete compression and refinement pipeline. 
The left column displays the original CoT annotated with our four discrete actions: 
\textsc{\textcolor{RoyalBlue}{Rewrite}} abstracts the core intent from conversational fillers (Step 0); 
\textsc{\textcolor{BurntOrange}{Fuse}} synthesizes fragmented arithmetic details into a single dense step (Steps 2--3); 
\textsc{\textcolor{BrickRed}{Prune}} eliminates redundant intermediate derivations from the chain (Steps 5--6); 
and \textsc{\textcolor{ForestGreen}{Keep}} preserves valid original steps without modification. 
The right column shows the transition from the compressed (high-density but disjointed) state to the final refined CoT (coherent and structured).
}
    \label{tab:full_case_study}
\end{table*}

\end{document}